\documentclass[runningheads]{llncs}
\usepackage{amsmath, amssymb, algorithmic, subfig}
\usepackage{graphicx, algorithm2e}
\usepackage[pagebackref=false,breaklinks=true,letterpaper=true,colorlinks,bookmarks=false]{hyperref}
\usepackage{array}
\usepackage{arydshln}
\usepackage{booktabs}
\usepackage{xcolor}
\newcommand{\todo}[1]{}
\renewcommand{\todo}[1]{{\color{red} \textbf{{#1}}}}

\newcommand{\toblue}[1]{{\color{black} {#1}}}

\begin{document}

\title{Peri-Diagnostic Decision Support Through Cost-Efficient Feature Acquisition at Test-Time}

\titlerunning{Peri-diagnostic decision support at test-time}  % abbreviated title (for running head)

\author{Gerome Vivar\inst{1}$^,$\inst{2}$^,$\thanks{G.V. and K.M. contributed equally to this work.} \and Kamilia Mullakaeva\inst{1}$^,$\inst{\star} \and Andreas Zwergal\inst{2} \and Nassir Navab\inst{1}$^,$\inst{3} \and Seyed-Ahmad Ahmadi\inst{1}$^,$\inst{2}
}
%index{Vivar, Gerome}
%index{Mullakaeva, Kamilia}
%index{Zwergal, Andreas}
%index{Navab, Nassir}
%index{Ahmadi, Seyed-Ahmad}

% %
\authorrunning{Vivar et al.} % abbreviated author list (for running head)

\institute{Technical University of Munich (TUM), Munich, GER\\
\and 
German Center for Vertigo and Balance Disorders (DSGZ), Ludwig-Maximilians-Universit{\"a}t (LMU), Munich, GER
\and
Whiting School of Engineering, Johns Hopkins University, Baltimore, USA}
\maketitle              % typeset the title of the contribution
\begin{abstract}
Computer-aided diagnosis (CADx) algorithms in medicine provide patient-specific decision support for physicians. These algorithms are usually applied after full acquisition of high-dimensional multimodal examination data, and often assume feature-completeness. This, however, is rarely the case due to examination costs, invasiveness, or a lack of indication. A sub-problem in CADx, which to our knowledge has not been addressed by the MICCAI community so far, is to guide the physician during the entire peri-diagnostic workflow, including the acquisition stage. We model the following question, asked from a physician's perspective: ``Given the evidence collected so far, which examination should I perform next, in order to achieve the most accurate and efficient diagnostic prediction?".
In this work, we propose a novel approach which is enticingly simple: use dropout at the input layer, and integrated gradients of the trained network at test-time to attribute feature importance dynamically. We validate and explain the effectiveness of our proposed approach using two public medical and two synthetic datasets. Results show that our proposed approach is more cost- and feature-efficient than prior approaches and achieves a higher overall accuracy. This directly translates to less unnecessary examinations for patients, and a quicker, less costly and more accurate decision support for the physician.

\end{abstract}

% %%%%%%%%%%%%%%%%%%%%%%%%%%%%%%%%%%%%%%%%%%%%%%%%%%%%%%%%%%%%%%%%%%%%%%%%%%%%%%%%%%%%%%%
% Introduction
% %%%%%%%%%%%%%%%%%%%%%%%%%%%%%%%%%%%%%%%%%%%%%%%%%%%%%%%%%%%%%%%%%%%%%%%%%%%%%%%%%%%%%%%
%
\section{Introduction}
\label{sec:intro}
The diagnostic workflow in medicine is ``an iterative process of information gathering, information integration and interpretation'' \cite{balog2015improvingdiagnosis}. Information is first acquired through a clinical history and interview, followed by alternating examinations and working diagnoses, until sufficient information has been aggregated for a final diagnosis. The decision which examination to perform next lies in the responsibility of the physician, who has to consider its medical indication, its invasiveness towards the patient, and often also its financial cost. Machine learning (ML) and computer-aided diagnosis (CADx) have a large potential for decision support in the clinic \cite{yanase2019cadxsurvey}. From a ML perspective, CADx is the task to learn the mapping of a multimodal feature vector onto a diagnostic label. Most CADx algorithms studied so far, however, ignore the acquisition stage, and provide decision support only at the end of the diagnostic workflow when all examination data is acquired and the feature vector is complete. As such, current CADx approaches miss out on the opportunity to aid the physician during the entire, \textit{peri-diagnostic workflow}, including the acquisition stage. In this work, we address this problem by i) iteratively suggesting the next most important examination/feature to acquire, while ii) considering the overall examination cost and aiming for a maximally accurate and efficient diagnostic prediction. To the best of our knowledge, the problem of \textit{peri-diagnostic decision support} has not been addressed in the MICCAI community so far. \\
\indent \textbf{Related works}: In ML literature, this problem is often described as budgeted or cost-sensitive feature acquisition. Most recent approaches can be roughly categorized into reinforcement learning (RL) and non-RL approaches. Among \textit{RL approaches}, \cite{shim2018joint} applied cost-sensitive n-step Q learning to CADx on Physionet (2012) and proprietary data. Kachuee et al. \cite{kachuee2019opportunistic} classify diabetes with Deep Q-networks (DQN) and Monte-Carlo dropout, and select the feature with the maximum confidence gain of the predictor network while considering cost. \cite{janisch2019classification} classify non-medical data with a DQN-variant that penalizes accumulated feature cost and incorrect predictions. RL-approaches have two important limitations: first, agent and predictor only work in tandem, neither has any utility or generalizability on its own. Second, unless agent and predictor are perfectly tuned, the network can quickly settle on a sub-optimal final classification accuracy in favor of low cost. Among the \textit{non-RL approaches}, \cite{contardo2016recurrent} classify fetal heartbeat patterns using Recurrent Neural Networks (RNN), which suggest the next feature through learned attention vectors as masks at every timestep. This can lead to suggesting several or repeated features at each timestep, and requires a fixed number of timesteps before its final prediction which can be inefficient. Kachuee et al. classify non-medical data \cite{kachuee2017context} and detect hypothyroidism \cite{kachuee2018dynamic} using denoising autoencoders (DAE). The DAE is trained with dropout at the input layer, and learns to reconstruct complete feature vectors from incomplete inputs. Next, the encoder part is fine-tuned and trained in tandem with a predictor network towards the final prediction task. At test time, the partial derivatives of all outputs with respect to each input feature are aggregated to form the total ``feature attribution''. In this context, it is important to note that feature attribution needs to fulfill four axioms, which have been derived in \cite{sundararajan2017axiomatic}. The gradient-based attribution only with respect to the input as performed in \cite{kachuee2017context,kachuee2018dynamic} violates the ``Sensitivity Axiom'' of feature attribution. This can lead to an acquisition of inefficient features \cite{sundararajan2017axiomatic} and ultimately, unnecessary patient examinations.\\ 
\indent \textbf{Contributions}: 1) We propose for the first time to apply Integrated Gradients (IG), an axiomatic feature attribution method, to the problem of dynamic, budgeted feature acquisition. 2) We propose Accumulated IG (AIG), for dynamically suggesting the next most important feature to acquire at test-time, and 3) we highlight the advantages of our proposed approach on two medical datasets and two explanatory datasets for illustration of its working principles.

%%%%%%%%%%%%%%%%%%%%%%%%%%%%%%%%%%%%%%%%%%%%%%%%%%%%%%%%%%%%%%%%%%%%%%%%%%%%%%%%%%%%%%%%
% Methods
%%%%%%%%%%%%%%%%%%%%%%%%%%%%%%%%%%%%%%%%%%%%%%%%%%%%%%%%%%%%%%%%%%%%%%%%%%%%%%%%%%%%%%%%
%
\section{Materials and Methods}
\label{sec:methods}
\subsection{Datasets and Preprocessing}
To evaluate our method, we utilized four datasets. The first two are publicly available medical datasets, to demonstrate the efficacy of our method on real-life data. The latter two datasets are non-medical, for further benchmarking as well as to illustrate the inner workings and limitations of the different methods we evaluate. For pre-processing, we perform outlier removal and scaling in NHANES and Thyroid (Winsorization of real-valued features to $[5,95]$-percentile, normalization to range [0,1]). \textbf{NHANES:} The ``National Health and Examination Survey" dataset \cite{cite_nhanes_website} contains demographic information, laboratory results, questionnaire, and physical examination data. The goal here is to predict diabetes (normal, pre-diabetes, and diabetes) based on measured fasting glucose levels. Costs for features were established in a crowd-sourcing approach \cite{kachuee2019syntheticdata}, and represent the total `inconvenience' of feature acquisition from a patient-perspective (including time burden, financial cost, discomfort, etc.) \cite{kachuee2019opportunistic}. The cost varies from 1 to 9 on a relative, numeric scale. We use all 92062 samples and 45 features in this dataset. \textbf{Thyroid:} The UCI Thyroid disease dataset \cite{quinlan87thyroid,dua2019ucirepository} poses a three-class classification problem (normal thyroid function vs. hyperfunction vs. or subnormal function). There are 21 features, representing demographic information, questionnaires and laboratory results that are important for thyroid disease classification. Feature costs are provided as part of the public dataset ``ann-thyroid" \cite{quinlan87thyroid}, and range from 1.00 to 22.78. We use all 7200 samples and 21 features.  \textbf{MNIST:} In the MNIST dataset \cite{lecun1998mnist}, we classify handwritten digit images in vectorized form, to simulate a tabular dataset. We use all 70,000 images with 784 features. We further assume a uniform cost of 1 for every pixel, to make our results comparable to related works. \textbf{Synthesized:} We also use a synthesized dataset as in \cite{kachuee2018dynamic}, to further explain and visualize the feature attribution process. The dataset consists of 16,000 samples with 64 dimensions. The first 32 dimensions contain salient information for classification, at a linearly increasing cost from 1 to 32. The second 32 dimensions contain no valuable information for classification, again at a linearly increasing cost of 1 to 32. Hence, intuitively, an efficient feature acquisition approach should choose only features from the first 32 dimensions. For a more detailed explanation, we refer the reader to \cite{kachuee2018dynamic}. 

\subsection{Problem Setting}
In this work, we consider the problem of patient-specific, dynamic feature acquisition at test time. The goal is to sequentially acquire features that can achieve the maximum prediction performance, as efficiently as possible. We aim for a model that is both cost- and feature-efficient, i.e. a model that achieves the maximum prediction performance with the least accumulated cost and smallest number of features possible. Formally, we consider the problem of predicting a target value $\hat{y} \in \operatorname{Y}$ based on a feature vector $\bar{x} \in \mathbf{R}^d$, which initially contains incomplete information about the patient at test-time. For clarity, we denote a complete feature vector as $x$ and an incomplete feature vector as $\bar{x}$.

\subsection{Efficient Feature Acquisition at test-time using Integrated Gradients.}
\label{methods:DAE_and_IG}
\begin{figure}
\begin{tabular}{cc}
  {\includegraphics[width = .6\textwidth]{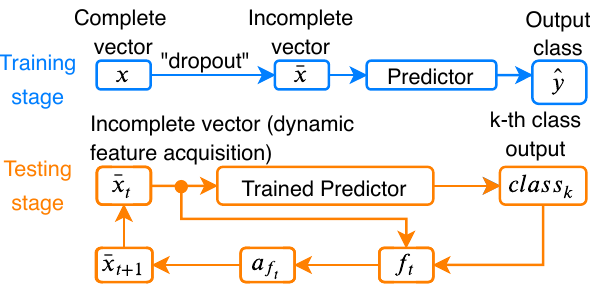}} &
  {\includegraphics[width = .4\textwidth]{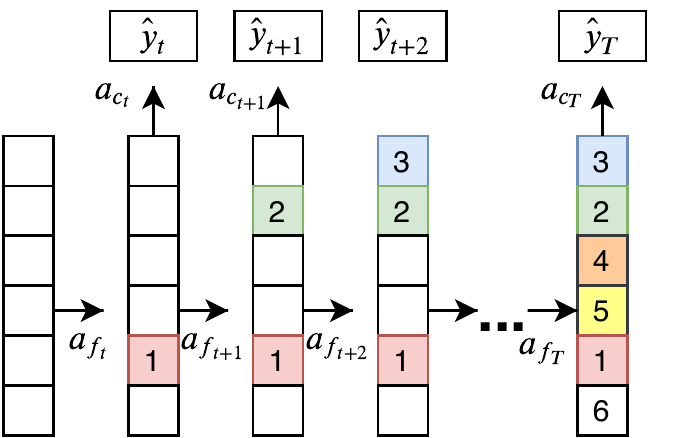}}
\end{tabular}
\caption{Illustration of our proposed network architecture (left panel) and an unrolled feature acquisition sequence at test-time (right panel).}
\label{fig:feature_acquisition_diag}
\end{figure}

To efficiently acquire features at test-time, we propose to use feature attribution by Integrated Gradients (IG) \cite{sundararajan2017axiomatic}. Previous works make use of backpropagation for feature acquisition \cite{kachuee2018dynamic,kachuee2017context}, by calculating the gradients of the network at the current input value. This, however, violates the  ``Sensitivity(a)'' axiom of feature attribution \cite{sundararajan2017axiomatic}, which states that if an input differs in one feature compared to a neutral baseline input, and if this leads to a different output, then that feature should be given a non-zero attribution. IG can be shown to uniquely satisfy the axiom ``Sensitivity(a)'', as well as the axiom ``Implementation Invariance'', which states that two different networks that produce the exact same outputs for the same inputs should produce the same feature attribution \cite{sundararajan2017axiomatic}. \\
Where previous gradient-based approaches \cite{kachuee2017context,kachuee2018dynamic} only take the gradient at the current input, IG takes a path integral of the gradients while linearly blending between a baseline input $x' \in \mathbf{R}^d$ and the actual input $x \in \mathbf{R}^d$, to avoid local gradients becoming saturated \cite{sundararajan2017axiomatic}. The baseline input $x'$ represents an ``absence'' of features and can be encoded as a zero-valued vector. Importantly, IG was originally designed for inference explanation, by computing feature attributions with respect to the known correct output class and model posterior. In our scenario, however, we do not know the output label of interest at test-time. We thus propose \textit{Accumulated IG (AIG)}, i.e. to aggregate the attributions of all input features from all possible output classes (see eqns. \ref{eq:IG} and \ref{eq:feature_attribution}). In addition, since we have an input $\bar{x}$ which is initially empty at test-time, \toblue{we have to use a
different baseline in order to be able to calculate AIG. We thus represent missing features with a neutral baseline at the central tendency (i.e. mean), analogous to mean-imputation in regular machine learning.} Here, accumulating the gradients implies combining attributions from $K$ different functions. This follows the ``Linearity Axiom'' of attribution theory, keeping AIG axiomatic as in the original IG formulation \cite{sundararajan2017axiomatic}. \\
\indent To handle missing information at test-time, previous works \cite{kachuee2018dynamic,kachuee2017context} proposed to use denoising autoencoders (DAE). We validate a combination of DAE with AIG in our experiments, but we also propose a simplified version without the need for auto-encoding. The simplified model is a vanilla multi-layer perceptron (MLP) trained end-to-end, while applying a Beta-distributed dropout layer to the input \cite{kachuee2018dynamic} to simulate missing information during training (see Fig. \ref{fig:feature_acquisition_diag}). \\
% using a low learning rate
%Now, we have the essential knowledge to discuss the implementation of IG.
\indent \textbf{Implementation Details:} 
We approximate the continuous IG as in \cite{sundararajan2017axiomatic} by a few discrete steps. We calculate the attribution along the $i$-th dimension with respect to one specific class ($k$) using:

\begin{equation}
    \operatorname{IG}_{i,k}^{approx} (\bar{x_i}, class_k) = (\bar{x}_i-{x}_i')\times \sum_{s=1}^m \frac{\partial F({x}'+\frac{s}{m}\times (\bar{x}-{x}'))}{\partial \bar{x}_i} \times \frac{1}{m}
    \label{eq:IG}
\end{equation}

where $\bar{x}$ is the input vector and ${x}'_i$ the baseline at the $i$-th dimension, $\frac{\partial{F}(.)}{\partial{\bar{x}_i}} $ is the partial derivative of the network's output with respect to input $\bar{x}_i$, and $m$ is the number of approximation steps of the path integral in IG. We then sum up all the attributions for the current feature from all classes and aggregate both positive and negative gradients. To account for cost-efficiency, we scale the attribution to a feature by the inverse feature cost:
\begin{equation}
f^{(i)}= \frac{| \sum_{k=1}^K \operatorname{IG}^{approx}(\bar{x}_i, class_k)|}{c_i}
\label{eq:feature_attribution}
\end{equation}

where  $f^{(i)}$ denotes the AIG feature attribute of input $\bar{x_i}$ and $c_i$ denotes its cost. Then $f_t \in \mathbf{R}^d$ is a vector which consists of AIG attributions of all features [$f^{(1)}_t$, $f^{(2)}_t$, ..., $f^{(d)}_t$] at timestep $t$. To determine which feature to acquire next, we take the index of the feature attribute with the maximum value: $a_{f_t}= \arg\max{(f_{t})}$, where $a_{f_t}$ denotes the feature to acquire at timestep $t$ as illustrated in \ref{fig:feature_acquisition_diag}. Using this newly acquired feature and previously acquired features we then perform classification ($a_{c_t}$) on this incomplete feature vector and obtain the label $y_t$. \toblue{We repeat this process until there are no more remaining features to acquire.} Alternatively, one can set a maximum allowed cost to constrain feature acquisition to a maximum allowed budget. The network architecture and an unrolled feature acquisition process are illustrated in Fig. \ref{fig:feature_acquisition_diag}.

%%%%%%%%%%%%%%%%%%%%%%%%%%%%%%%%%%%%%%%%%%%%%%%%%%%%%%%%%%%%%%%%%%%%%%%%%%%%%%%%%%%%%%%%
% Results
%%%%%%%%%%%%%%%%%%%%%%%%%%%%%%%%%%%%%%%%%%%%%%%%%%%%%%%%%%%%%%%%%%%%%%%%%%%%%%%%%%%%%%%%
%
\section{Results and Discussion}
\label{sec:results}

\subsection{Experimental setup}
\indent \textbf{Baseline model comparison:} We evaluate our work against several baseline and state-of-the-art approaches in budgeted feature acquisition, including a recent deep-RL \cite{kachuee2019opportunistic}, two non-RL \cite{kachuee2018dynamic,kachuee2017context}, and a random feature selection based approaches. We split the datasets into 15\% test set and 85\% training set, where 15\% of the latter is used for validation. We use Adam optimization \cite{kingma2014adam} implemented in PyTorch \cite{paszke2019pytorch} on a single-GPU workstation (Nvidia GTX 1080 Ti). For RL, we used the author implementation and parametrizations of Opportunistic Learning \cite{kachuee2019opportunistic}, to train an agent for 11,000 episodes. For comparison, all methods including our own are based on a two-layer multilayer perceptron (MLP) [64 and 32 units]. The non-RL approaches we compare against are Dynamic Feature Query (DPFQ) \cite{kachuee2017context} and Feature Acquisition Considering Cost at Test-time (FACT) \cite{kachuee2018dynamic}. Again, we use the same MLP architecture for the encoder [64, 32], decoder [32, 64], and predictor [32, 16, $K$ classes]. For the binary layer in FACT, we use the identical 8-bit representation as in \cite{kachuee2018dynamic}. Further, to randomly drop entries, we use a Beta-distribution with $\alpha = 1.5$ and $\beta = 1.5$, following \cite{kachuee2018dynamic}. \\
\textbf{Proposed:} We use a vanilla MLP (encoder [64, 32], and predictor [32, 16, $K$ classes]). We used the Adam optimizer in PyTorch with a low learning rate ($lr = 1e-4$). We use $m = 50$  for the number of steps in the integral approximation in eqn. \eqref{eq:IG}.

\subsection{Feature acquisition performance}
\begin{figure}
\begin{tabular}{cc}
{\includegraphics[width = 0.5\textwidth]{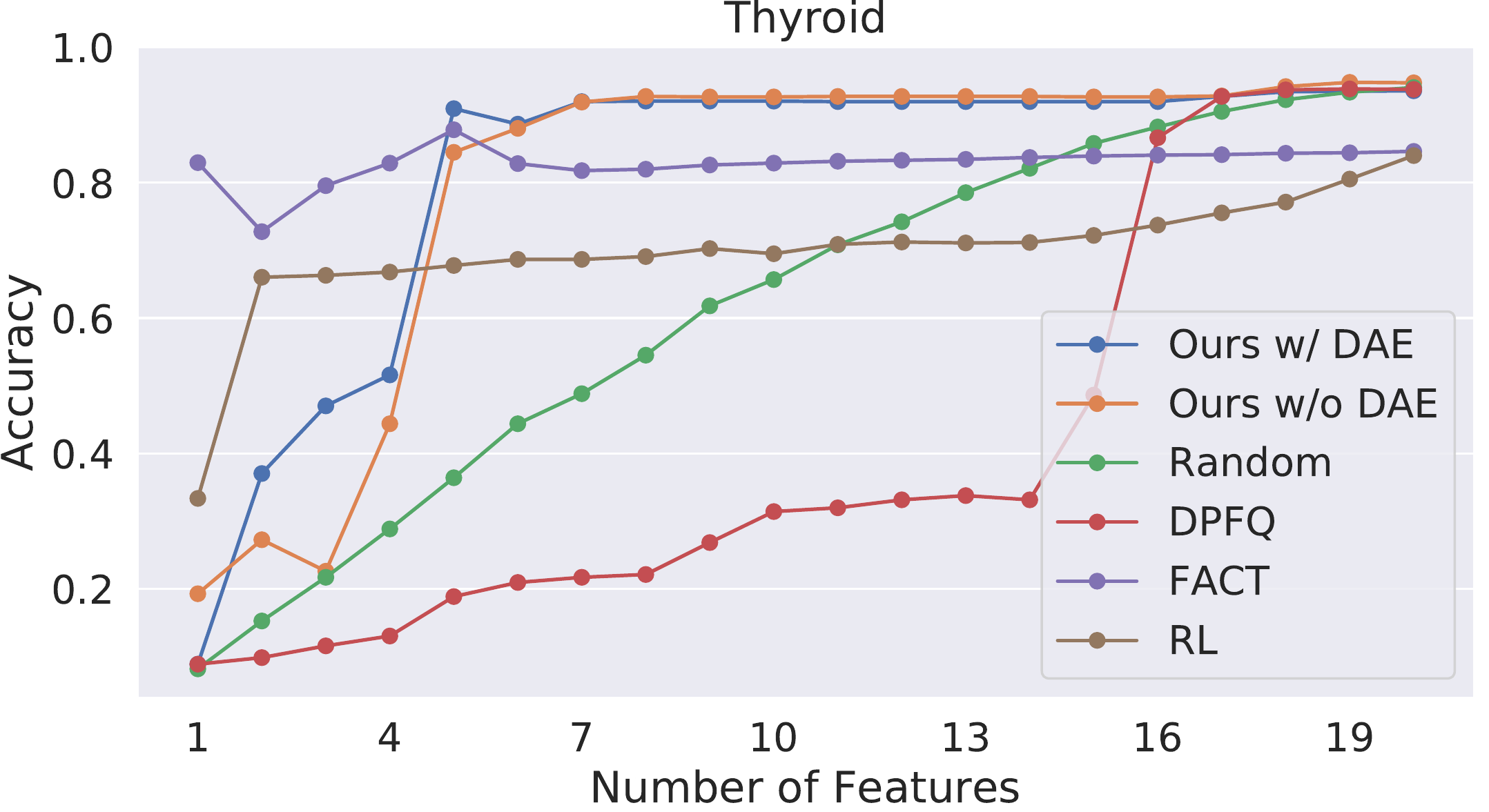}} &
{\includegraphics[width = 0.5\textwidth]{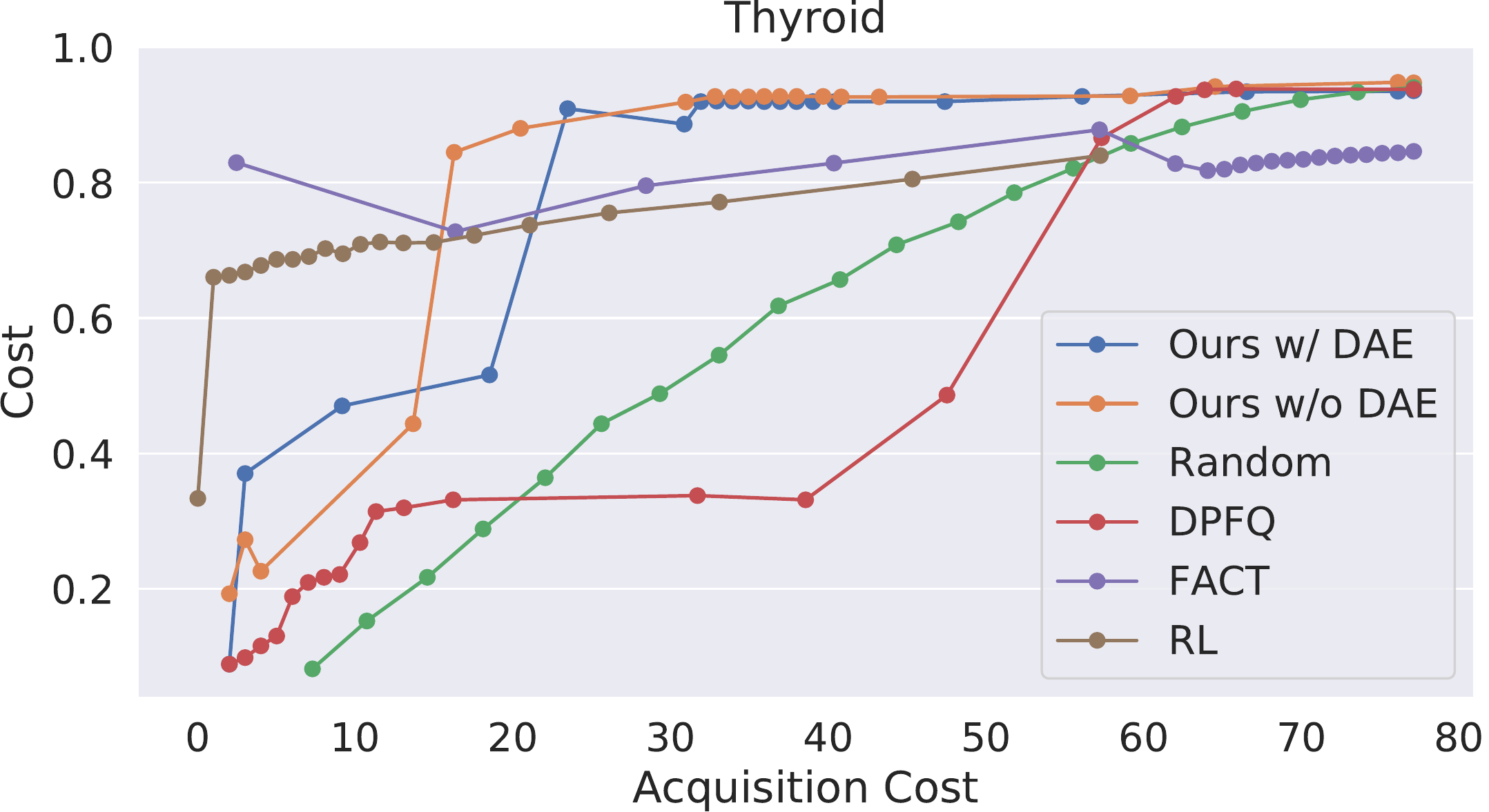}} \\
{\includegraphics[width = 0.5\textwidth]{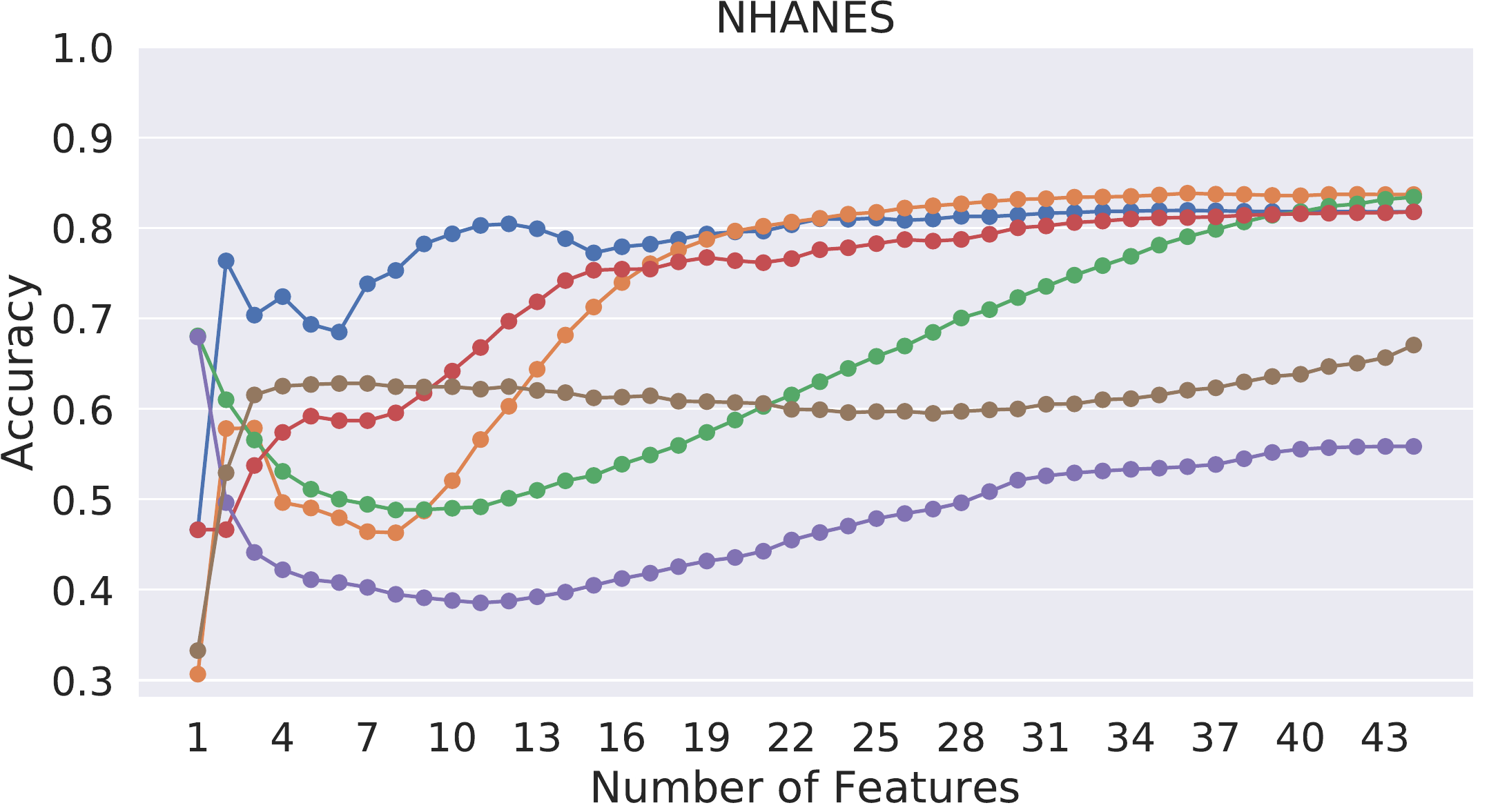}} &
{\includegraphics[width = 0.5\textwidth]{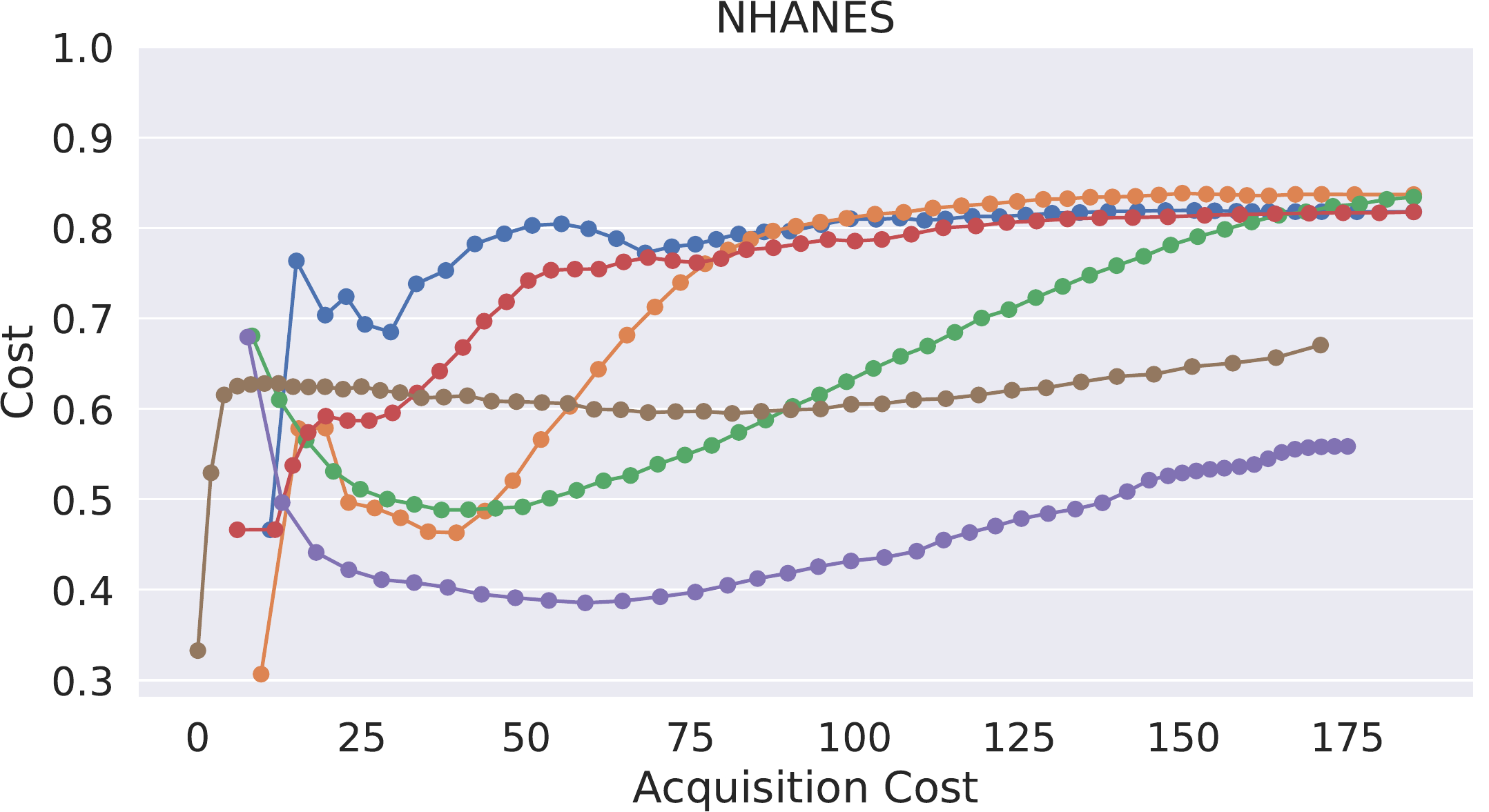}} \\
{\includegraphics[width = 0.5\textwidth]{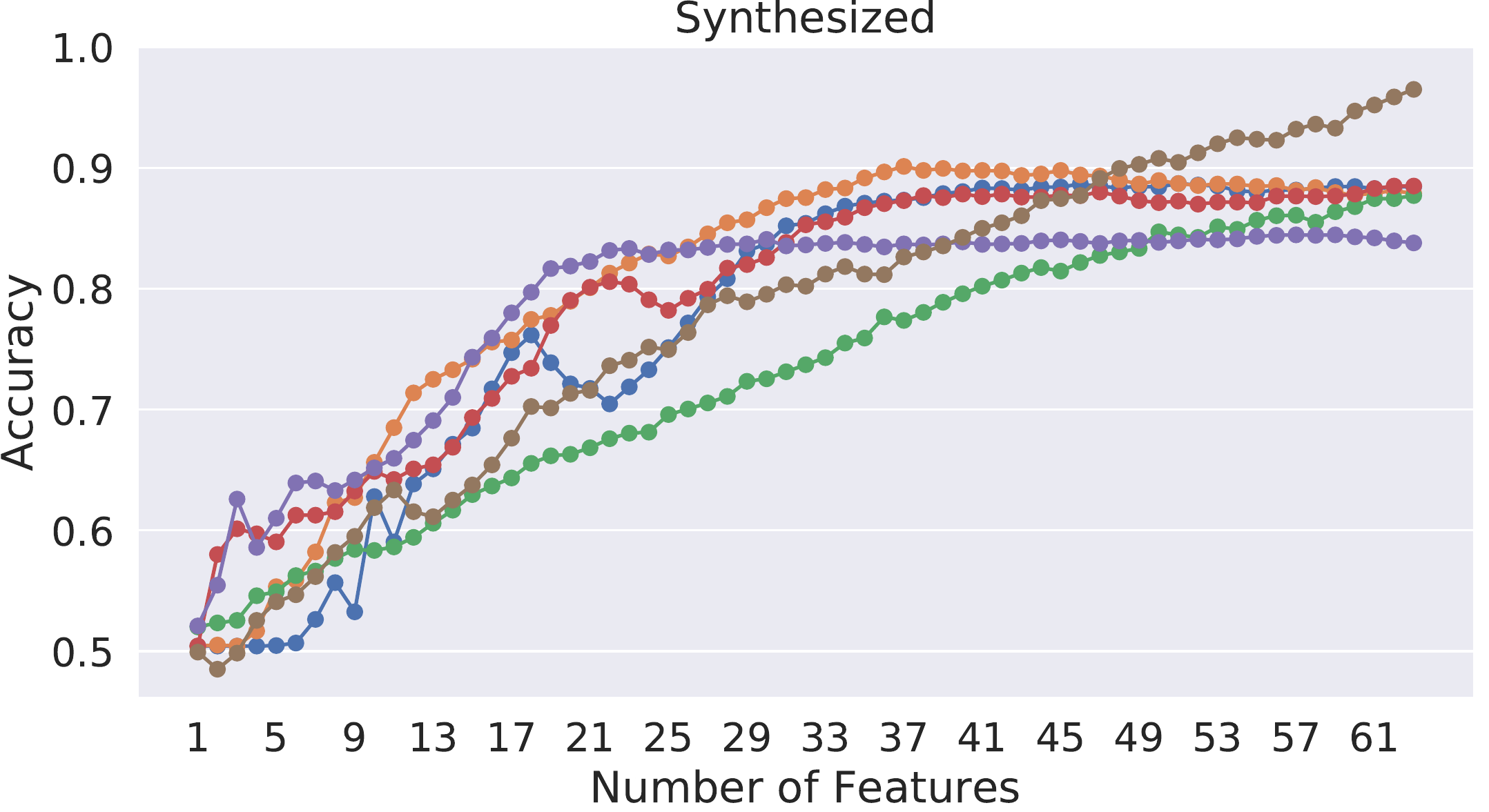}} &
{\includegraphics[width = 0.5\textwidth]{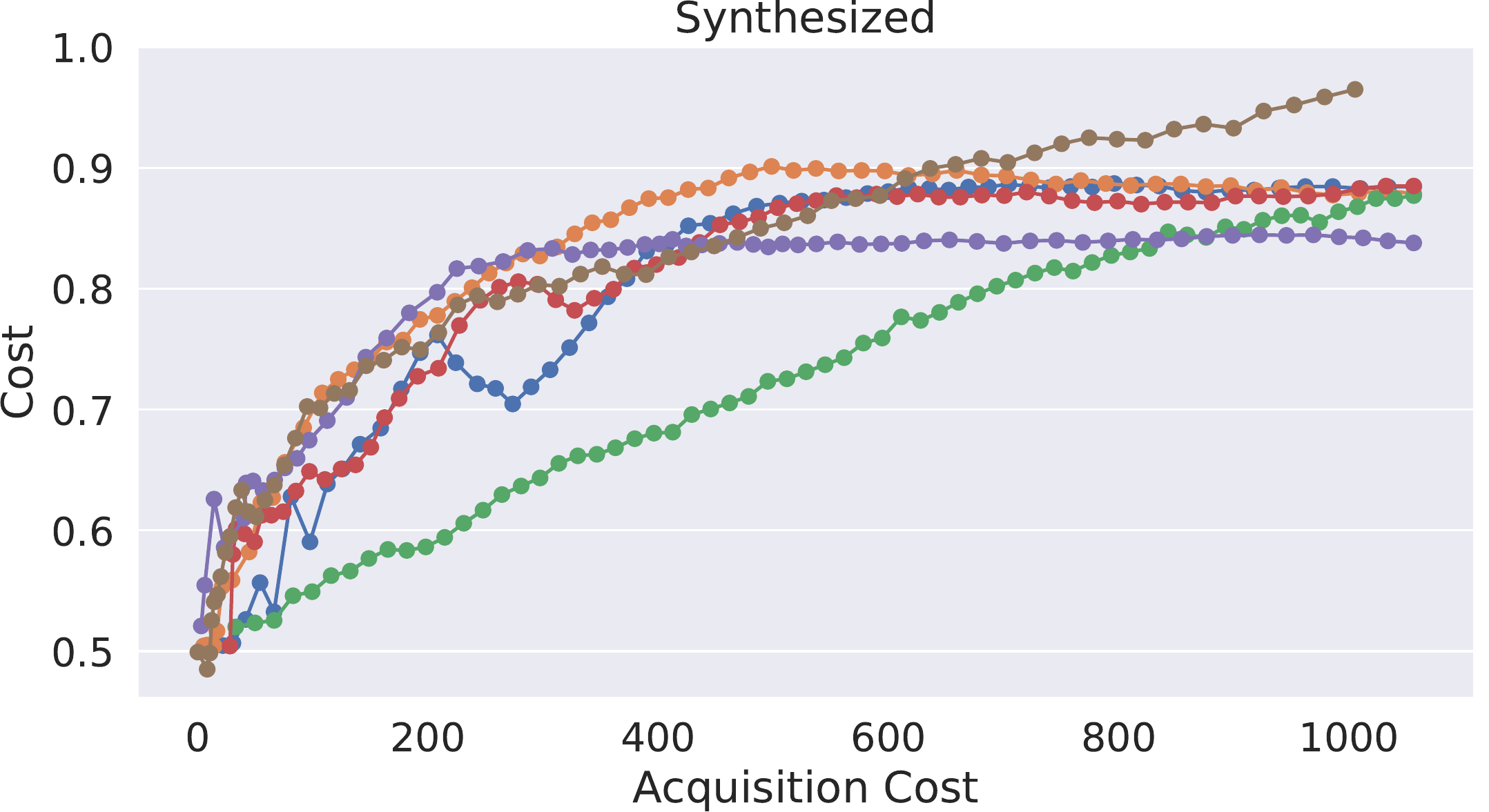}} \\
\multicolumn{2}{c}{ {\includegraphics[width = 0.5\textwidth]{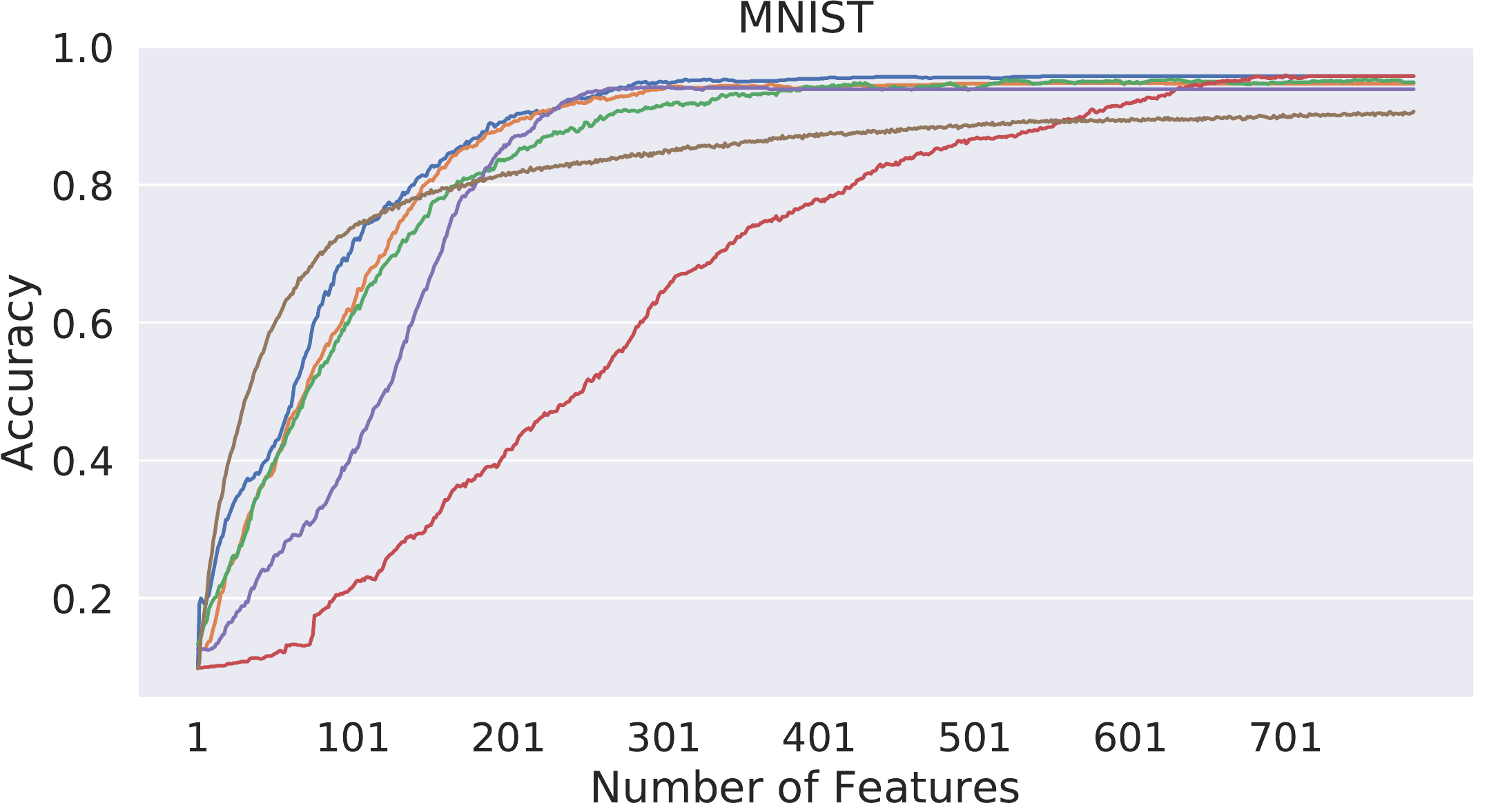}} }
 \\
\end{tabular}
\caption{Comparison of feature aggregation methods against our proposed approach, on four datasets (row 1: Thyroid; row 2: NHANES; row 3: Synthesized; row 4: MNIST with uniform feature cost). Left column: Feature count vs. accuracy curves; right column: Accumulated feature cost vs. accuracy. The compared baseline approaches denote: Random (random feature selection), DPFQ \cite{kachuee2017context}, FACT \cite{kachuee2018dynamic}, RL \cite{kachuee2019opportunistic}. Our approach is consistently most feature- and cost-efficient and achieves the highest classification final accuracy.}
\label{fig:baseline}
\end{figure}

We compare our work with previous deep-RL \cite{kachuee2019opportunistic} and non-RL \cite{kachuee2018dynamic,kachuee2017context} techniques to evaluate the effectiveness of our proposed method. \toblue{We observe that our proposed AIG approach with and without DAE outperforms the SOTA methods, with a particularly large margin in the two medical datasets.} Overall, our approach is the most cost- and feature-efficient (see Fig. \ref{fig:baseline}) and consistently achieves the highest overall classification accuracy. The only exception is RL for Synthesized data, but otherwise RL lacks robustness. Our method's feature-efficiency is evident e.g. \toblue{in Thyroid and NHANES, on average, it is able to outperform the SOTA and reach the maximum classification accuracy after just 7  ($\sim 33\%$) and 10 ($\sim 22\%$) features, respectively (see Fig. \ref{fig:baseline}).} Importantly, this directly translates to the avoidance of unnecessary examinations and a much faster time-to-diagnosis, without requiring patients to undergo all examinations. Apart from feature-efficiency, our approach is also cost-efficient, e.g. spending only \toblue{$\sim 20$ ($\sim 25\%$)}  units of cost in Thyroid, and \toblue{$\sim 50$ ($\sim 29\%$)} units in NHANES to achieve maximum classification performance. Further, methods like RL or DPFQ may choose cheaper features first, despite little gain in classification accuracy (see Fig. \ref{fig:baseline}, right Thyroid panel), whereas our method suggests more costly features in the beginning, at the benefit of reaching the highest classification accuracy almost instantly.

\subsection{Interpretation of patient-specific feature acquisition}
\label{sec:Results_heatmap_interpretation}
\begin{figure}
\begin{tabular}{ccccc}
Ours w/ DAE & Random & DPFQ & FACT & RL \\
% Thyroid
%\hdashline
{\includegraphics[width = 0.19\textwidth]{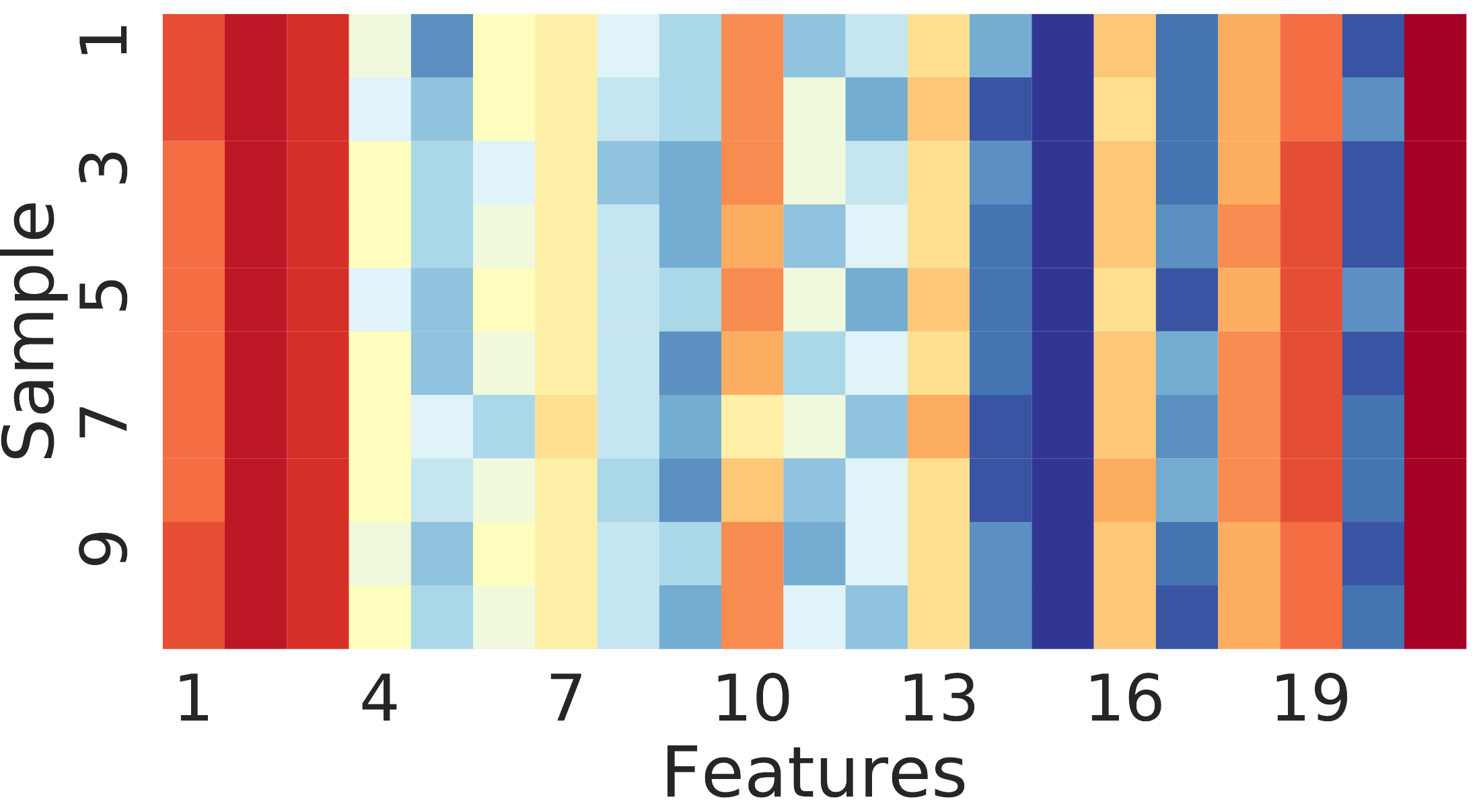}} &
{\includegraphics[width = 0.19\textwidth]{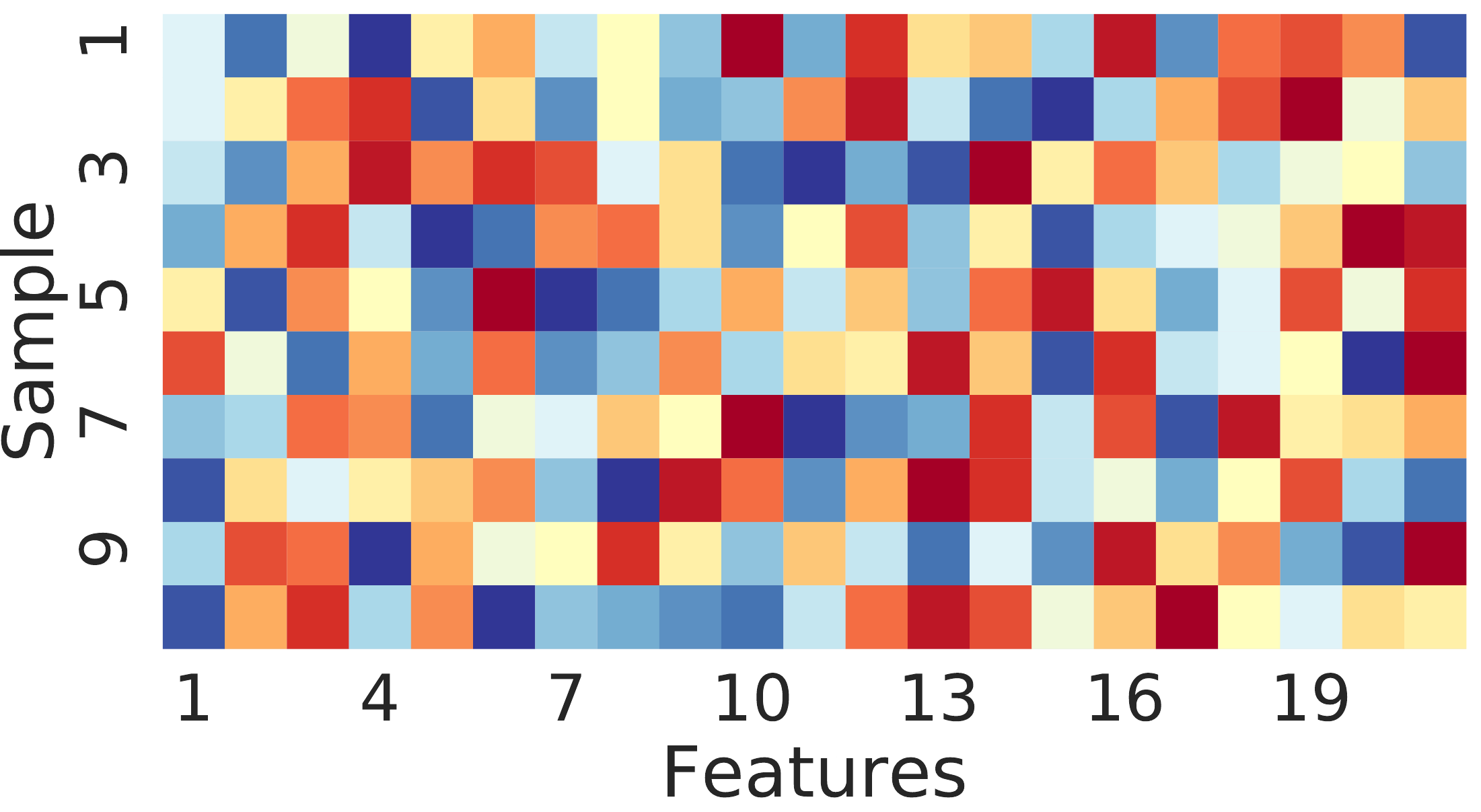}} &
{\includegraphics[width = 0.19\textwidth]{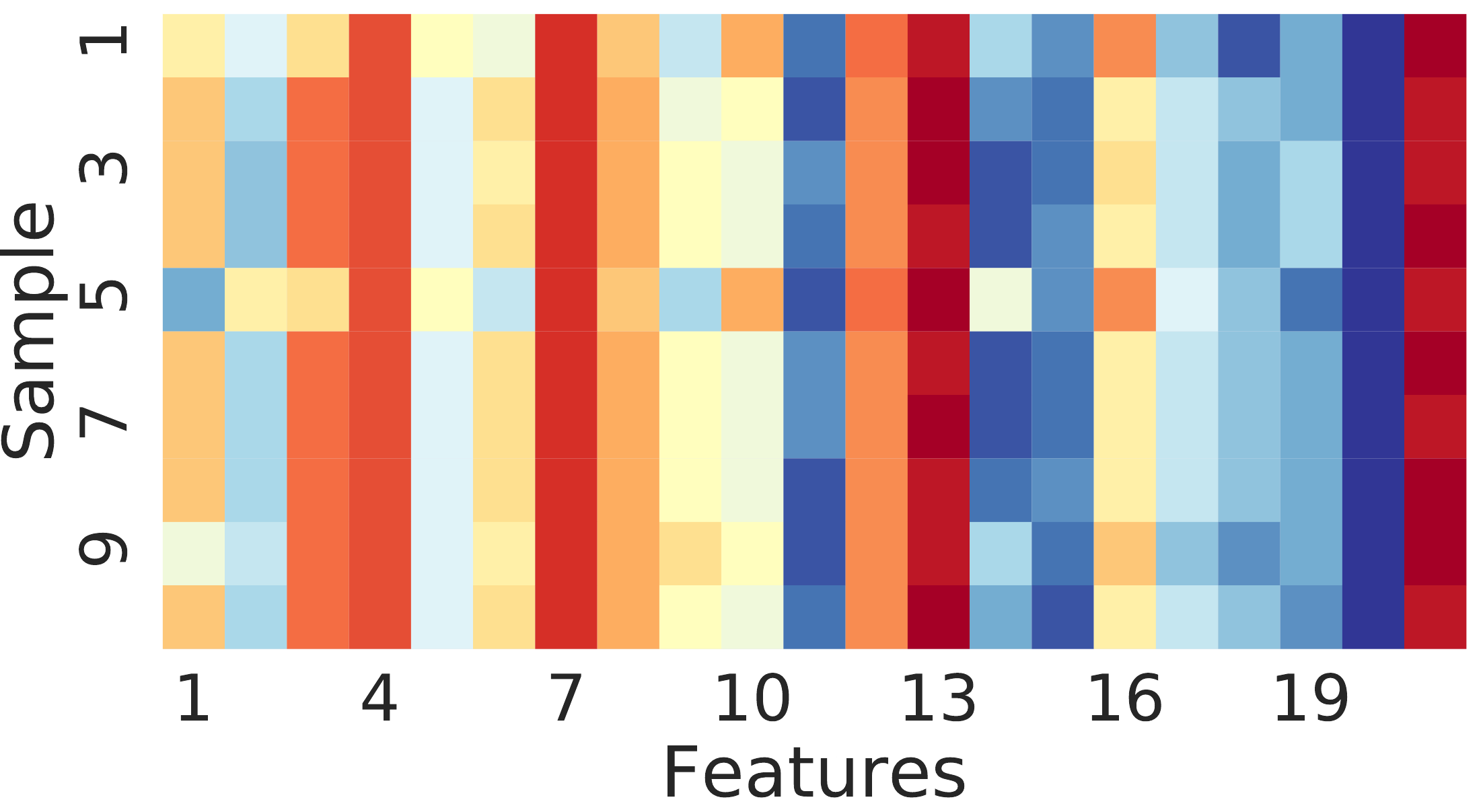}} &
{\includegraphics[width = 0.19\textwidth]{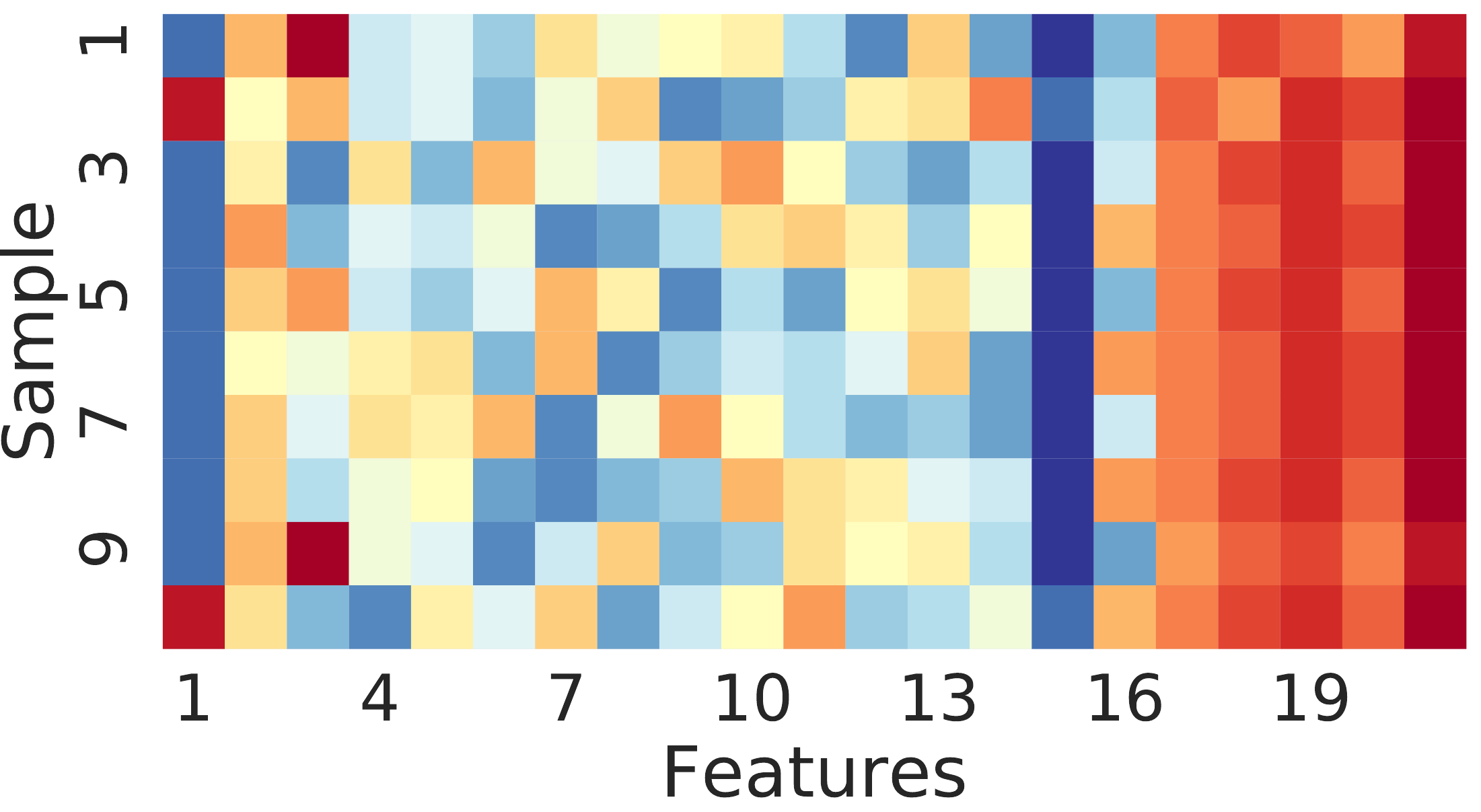}} &
{\includegraphics[width = 0.19\textwidth]{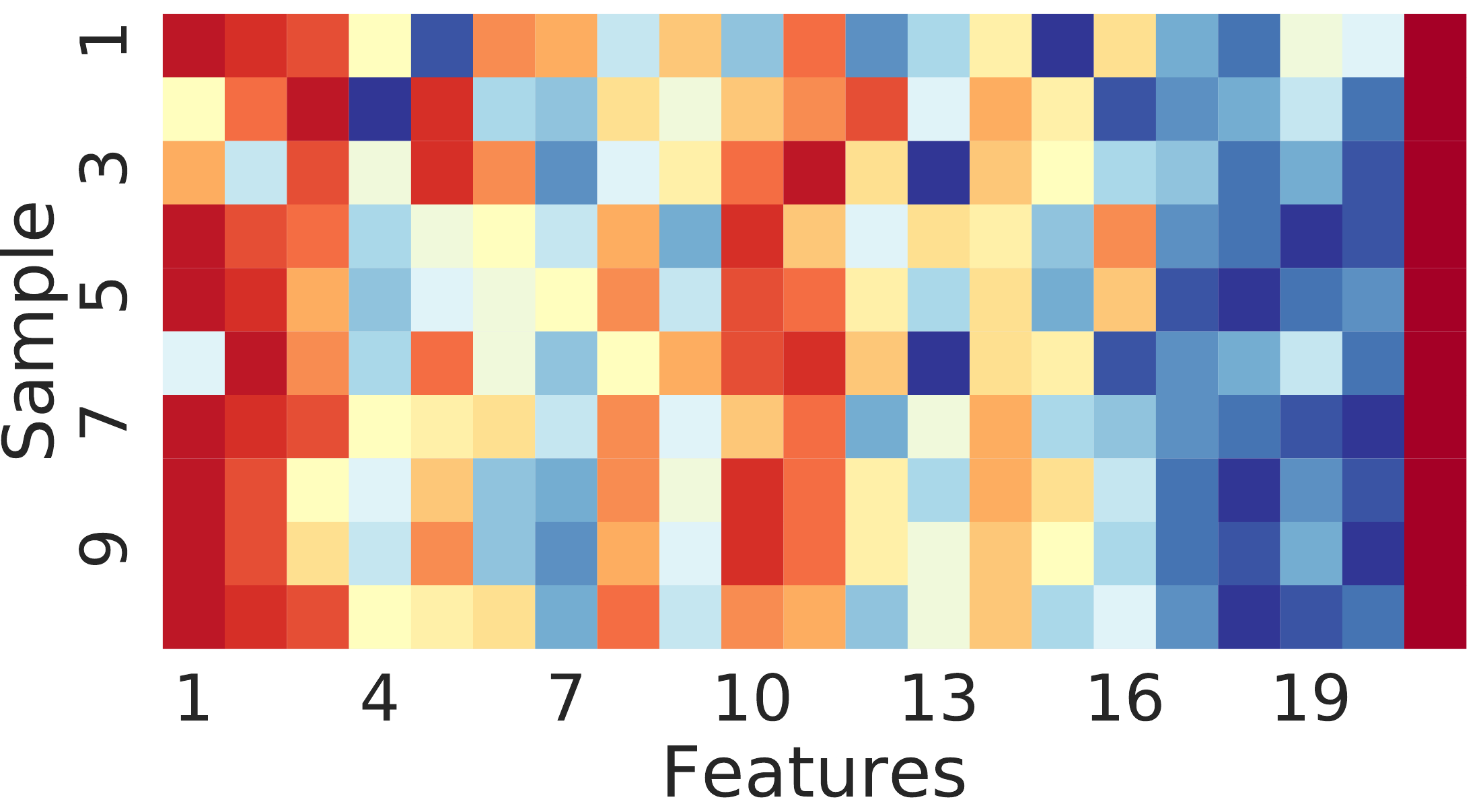}} \\
%\hdashline
% % NHANES
{\includegraphics[width = 0.19\textwidth]{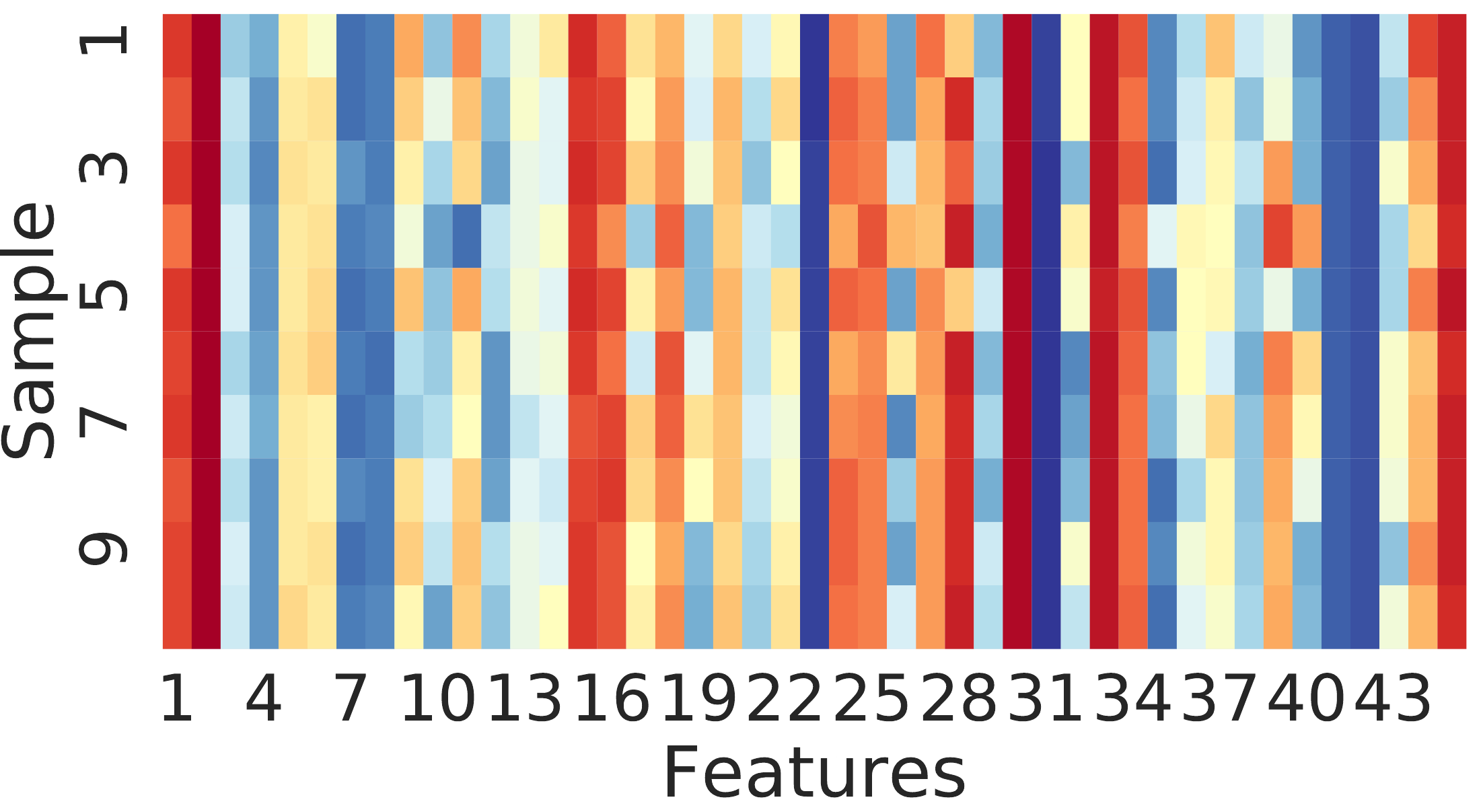}} &
{\includegraphics[width = 0.19\textwidth]{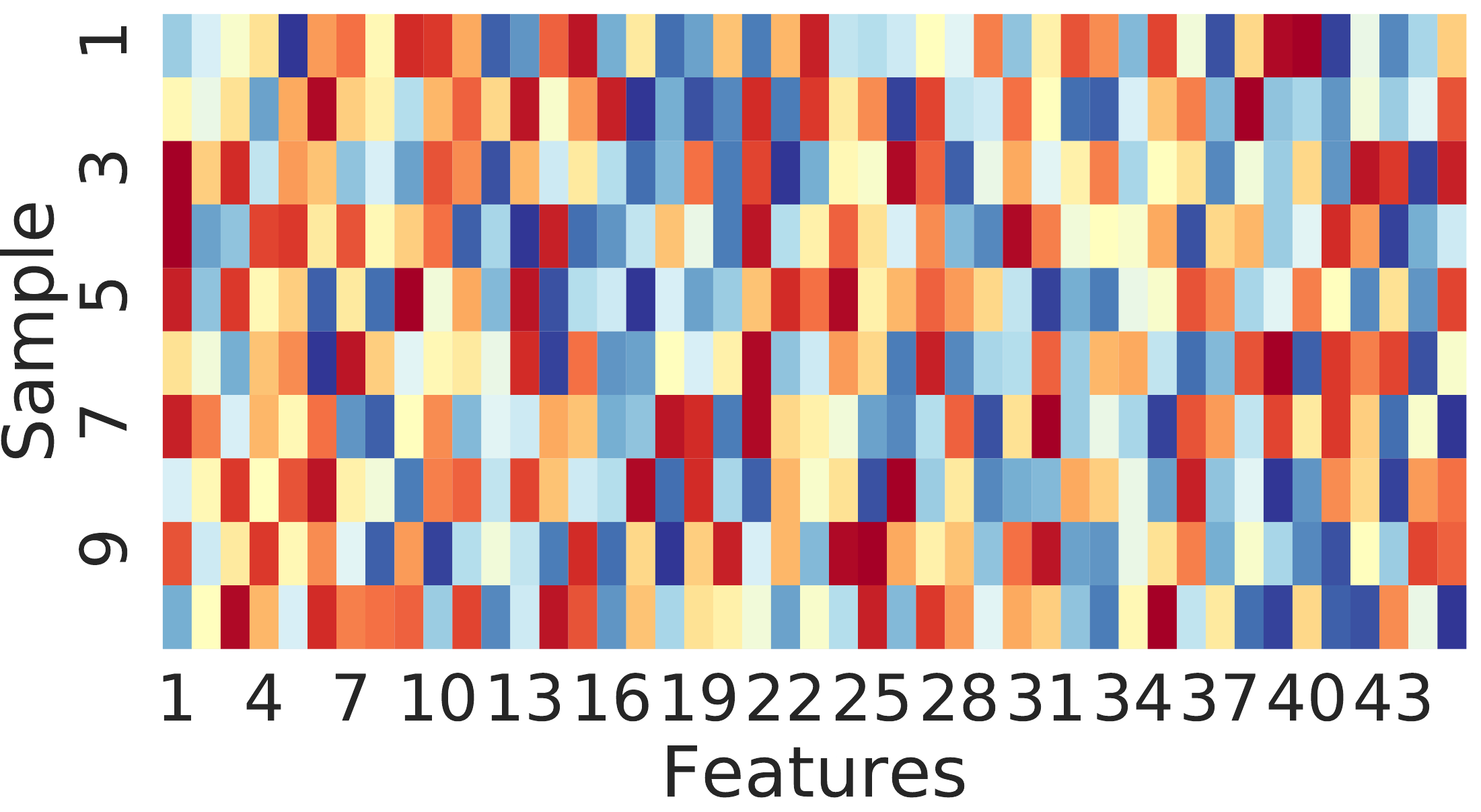}} &
{\includegraphics[width = 0.19\textwidth]{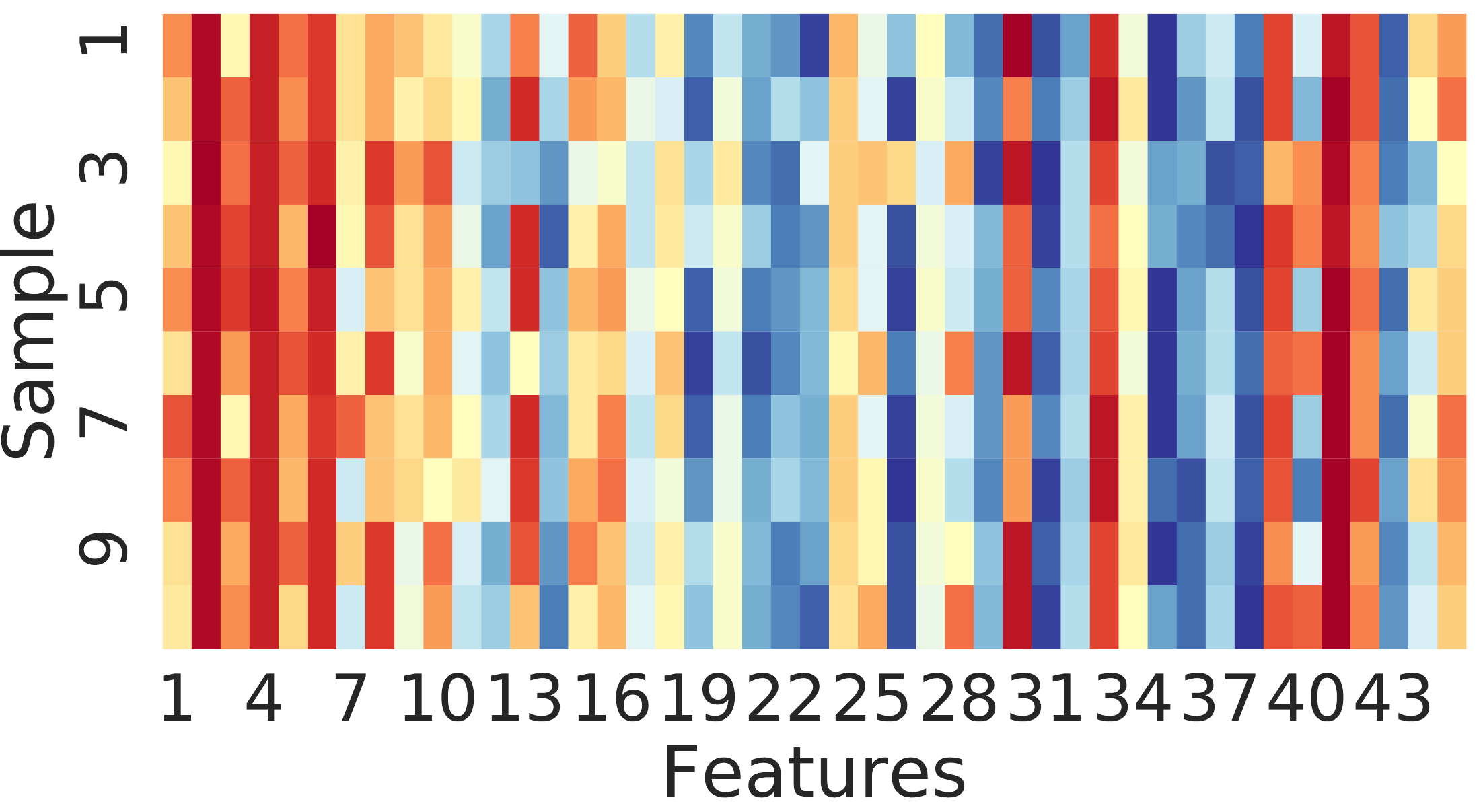}} &
{\includegraphics[width = 0.19\textwidth]{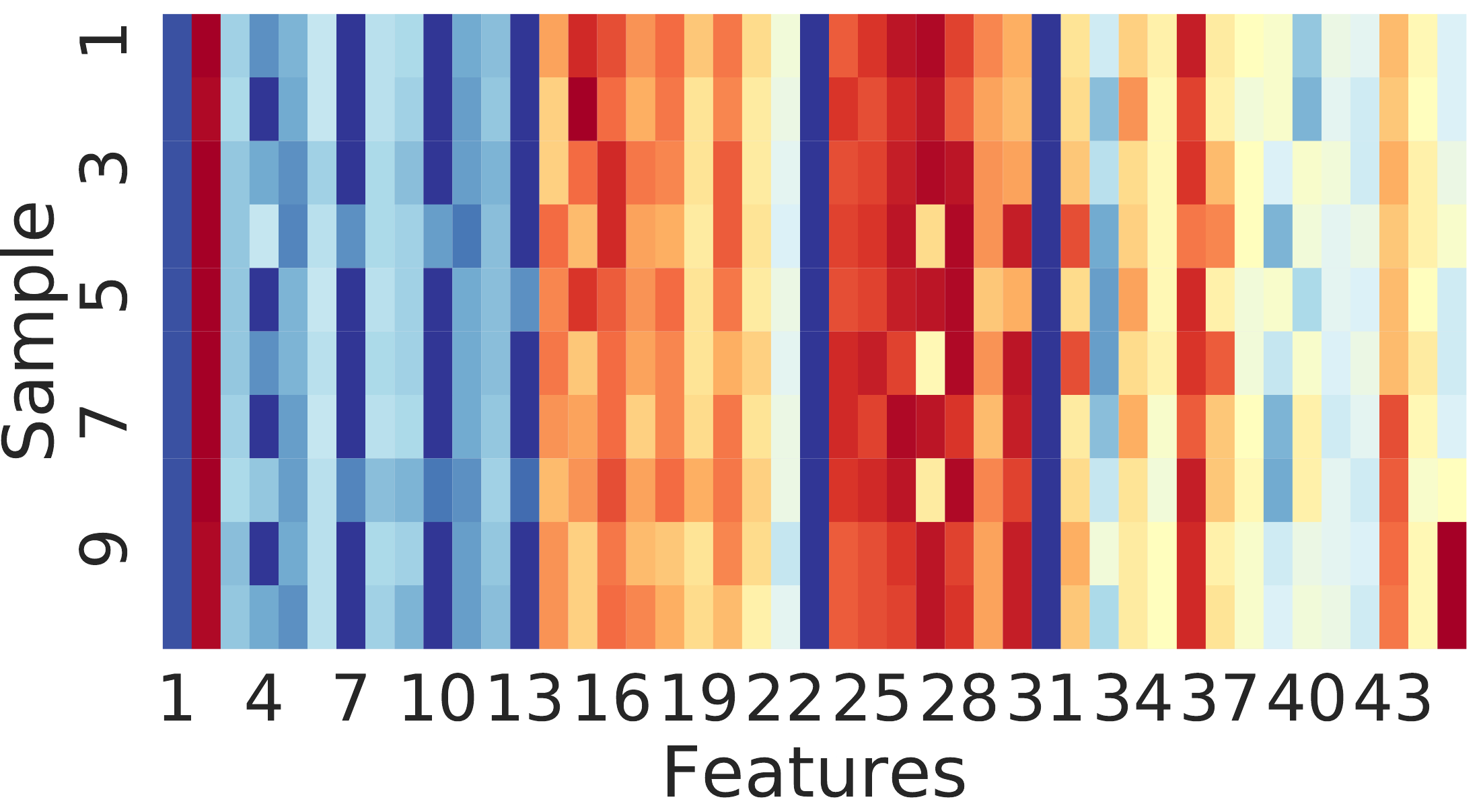}} &
{\includegraphics[width = 0.19\textwidth]{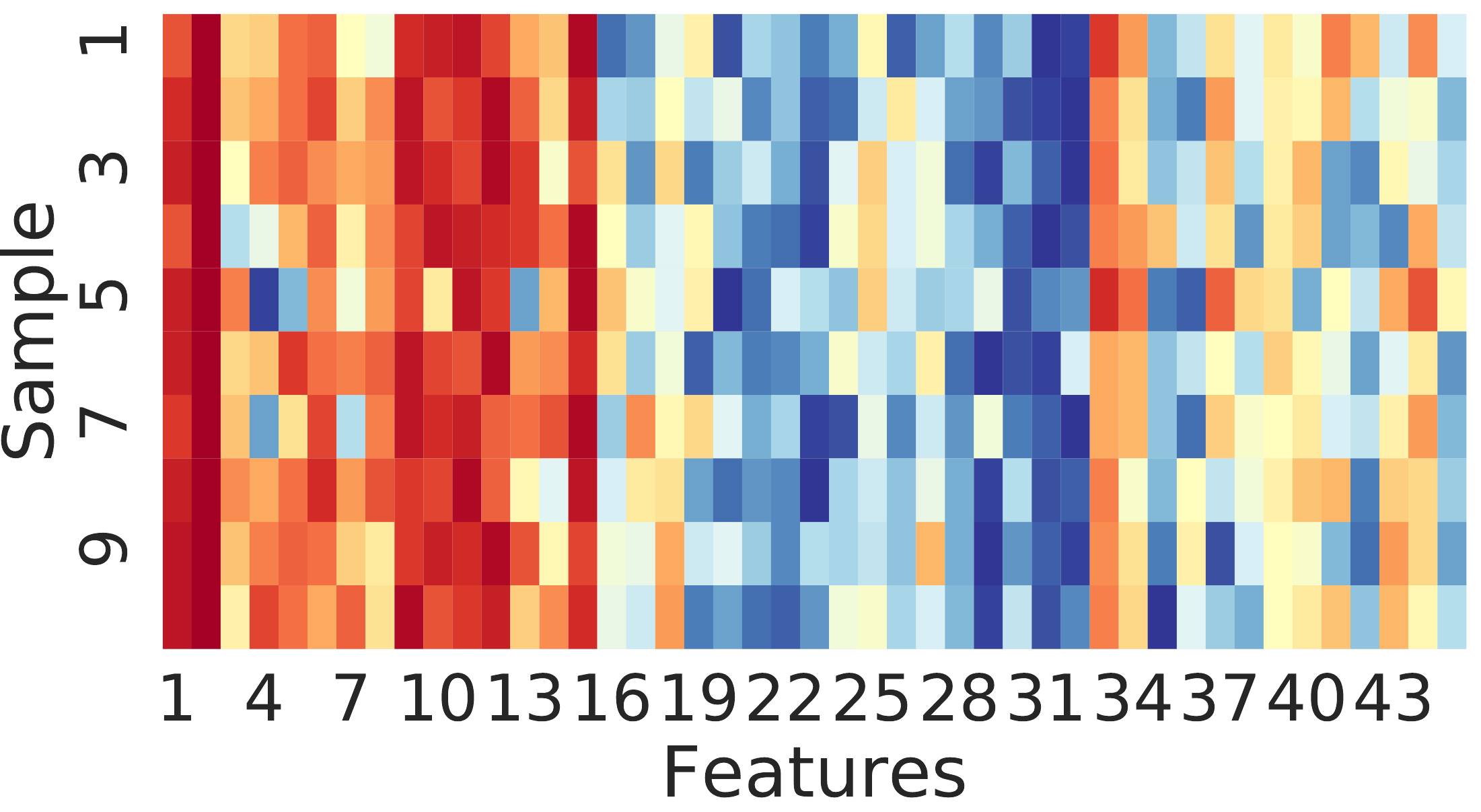}} \\
% Synthesized
{\includegraphics[width = 0.19\textwidth]{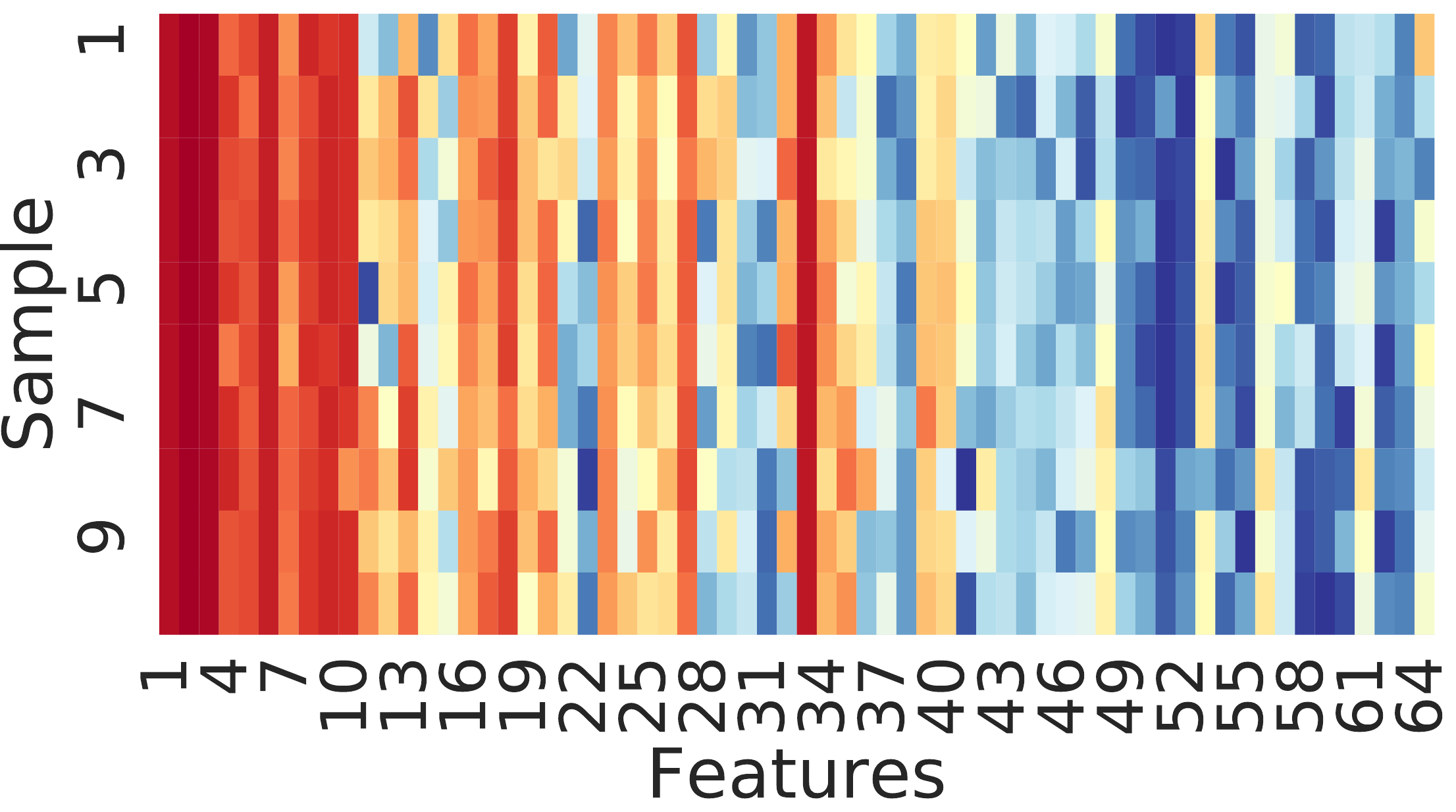}} &
{\includegraphics[width = 0.19\textwidth]{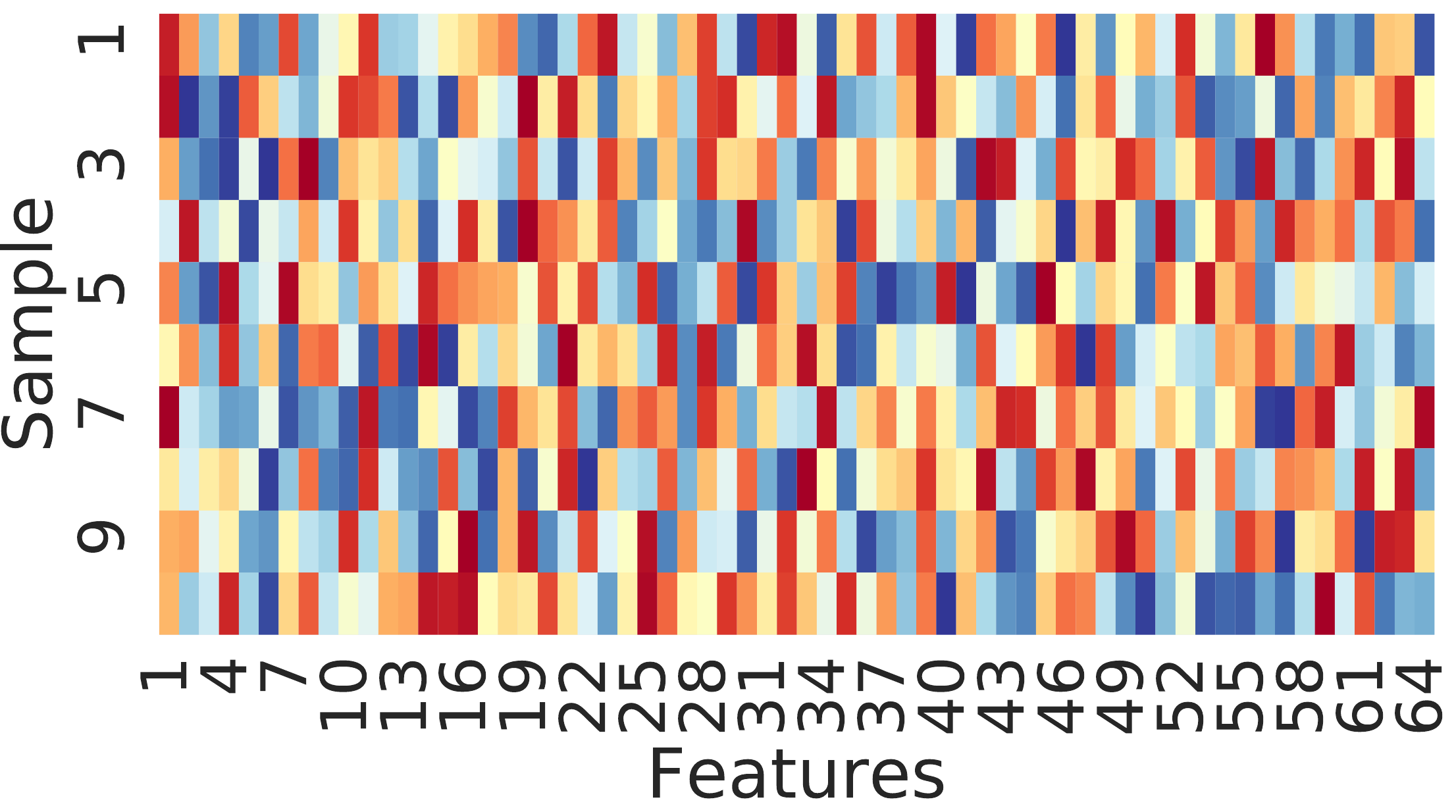}} &
{\includegraphics[width = 0.19\textwidth]{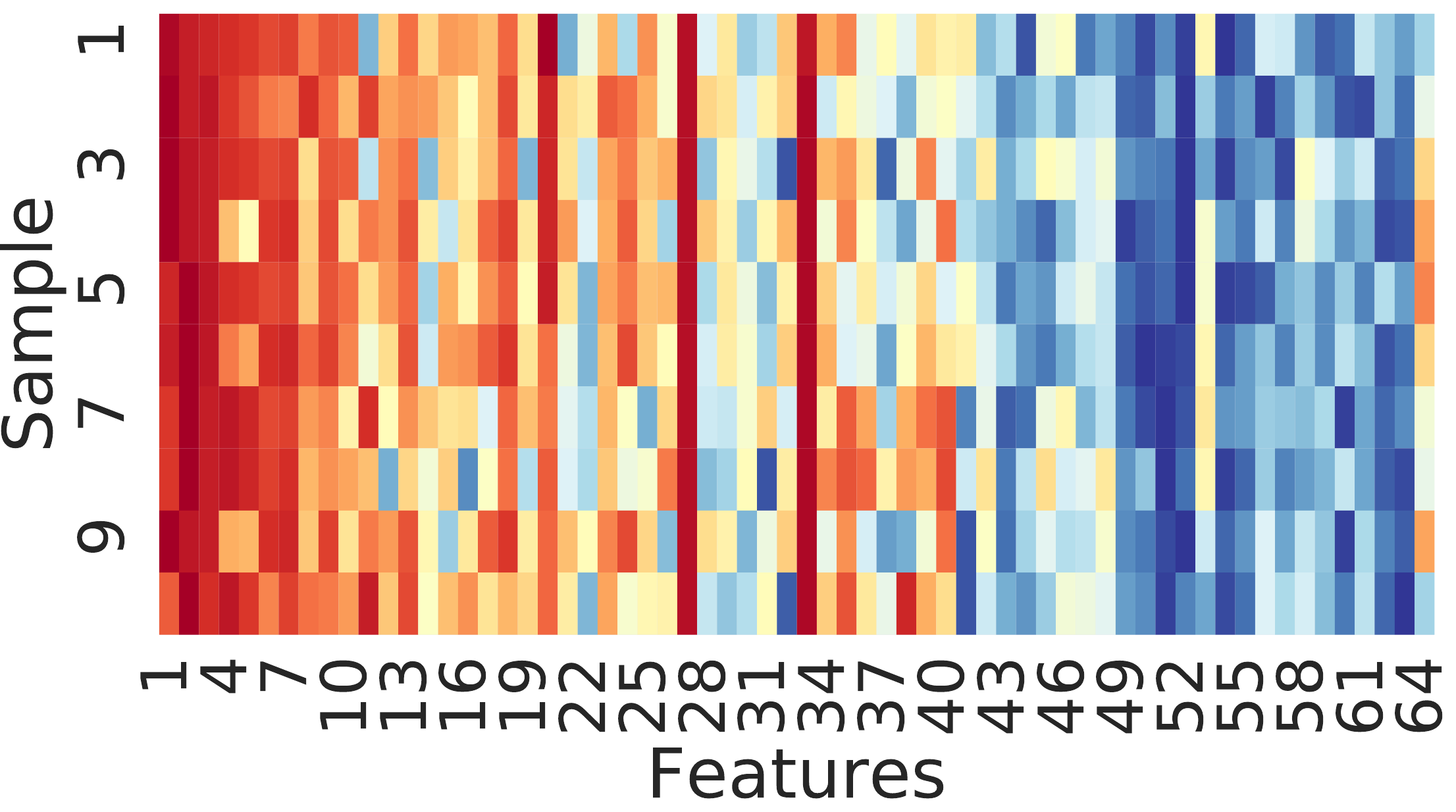}} &
{\includegraphics[width = 0.19\textwidth]{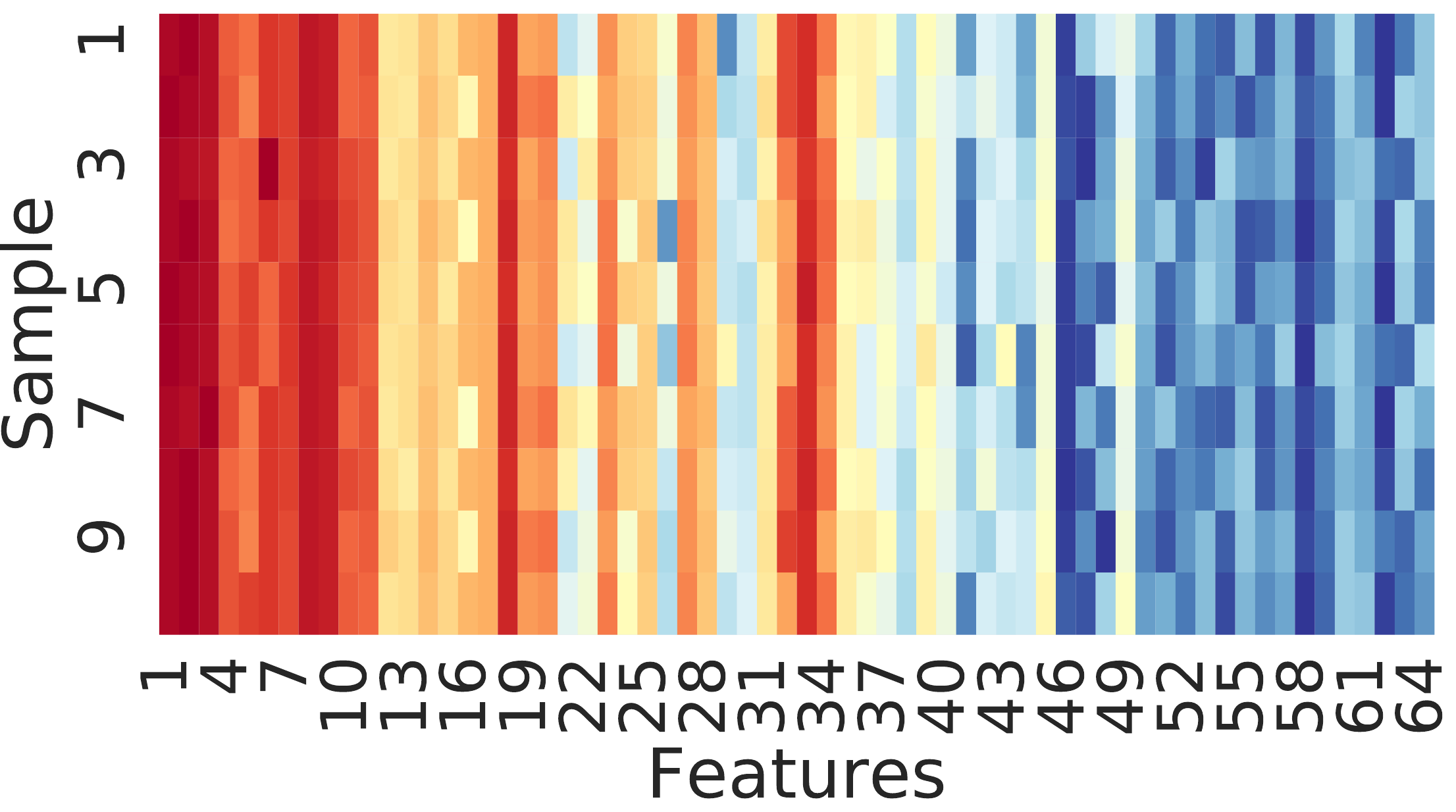}} &
{\includegraphics[width = 0.19\textwidth]{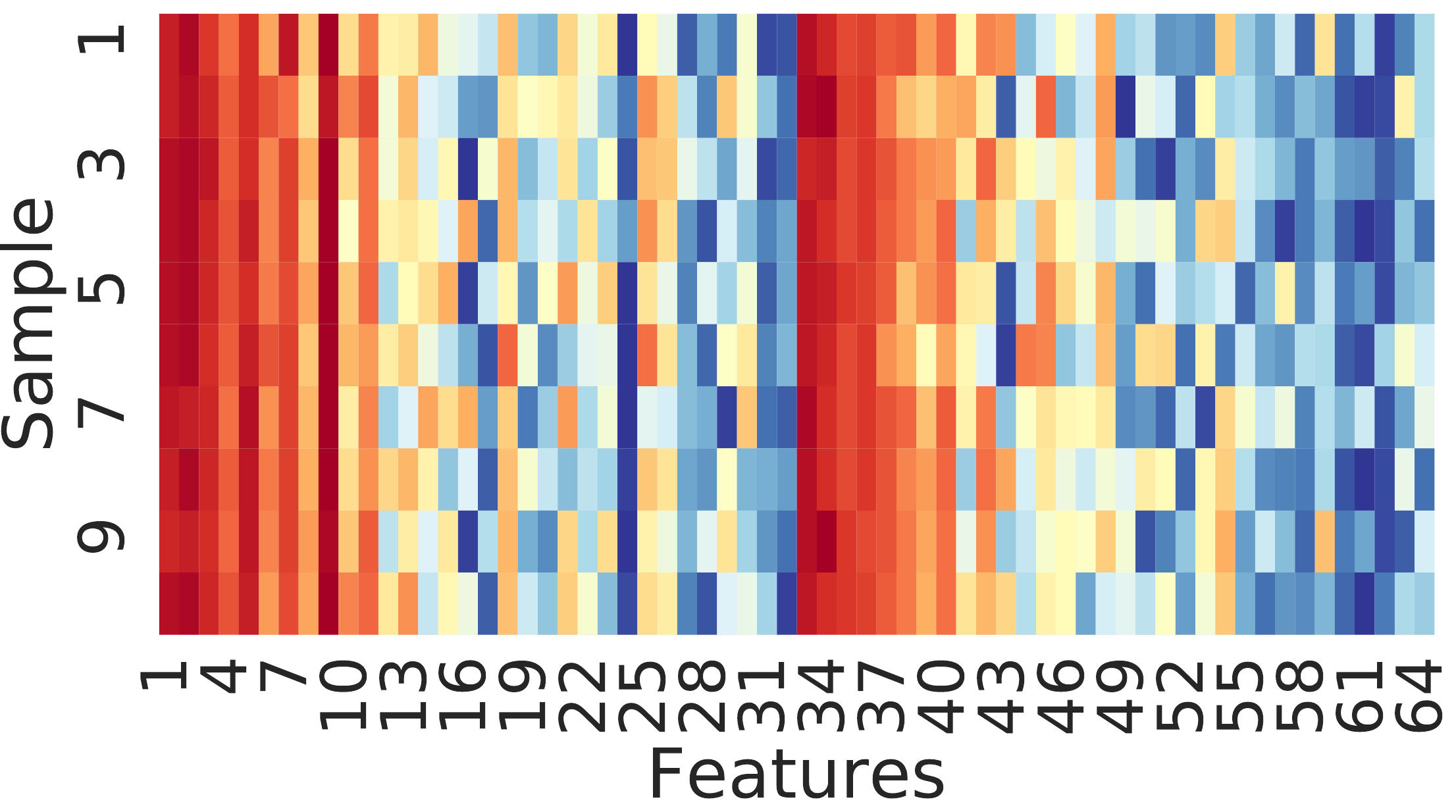}} \\
%\hdashline
% MNIST
{\includegraphics[width = 0.185\textwidth]{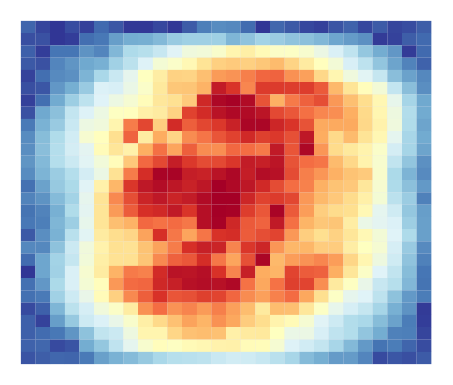}} &
{\includegraphics[width = 0.185\textwidth]{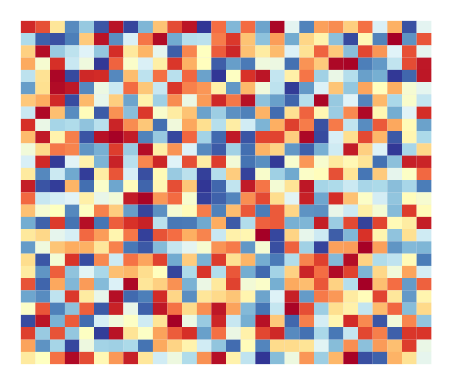}} &
{\includegraphics[width = 0.185\textwidth]{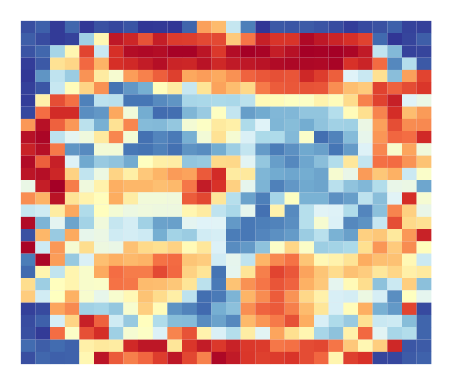}} &
{\includegraphics[width = 0.185\textwidth]{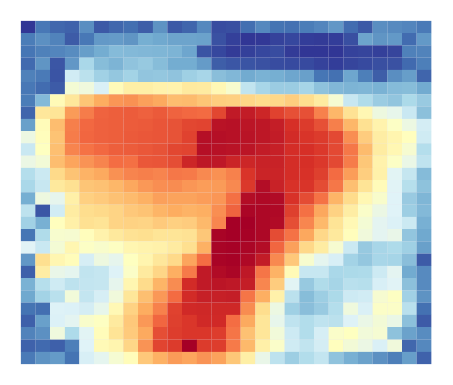}} &
{\includegraphics[width = 0.185\textwidth]{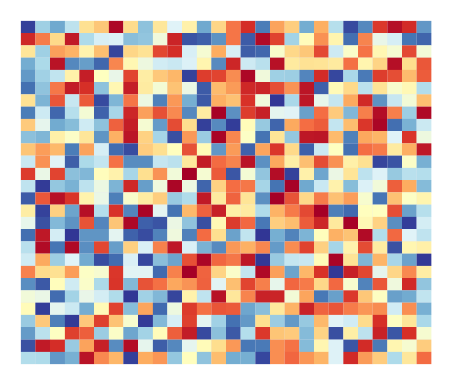}} \\
%\hdashline
\end{tabular}
\caption{Feature acquisition heatmaps for all datasets. From top to bottom: UCI-Thyroid (10 patient samples), NHANES (10 patient samples), Synthesized (10 samples), and MNIST. Warmer colors denote higher priority for feature acquisition.}
\label{fig:heatmap}
\end{figure}

We also use test samples of each dataset to visualize and discuss the order of feature selection by the different methods. We show heatmaps in Fig. \ref{fig:heatmap}, where warmer colors denote higher priority in the feature acquisition. We plot ten test samples for datasets Thyroid, NHANES and Synthesized, and one test image from MNIST.
%\todo{(more examples in the supplementary material)}. 
In Fig. \ref{fig:heatmap}, we observe that our proposed approach \toblue{ initially always acquires the most informative feature, before starting to acquire features in an instance- or patient-specific manner. For example, in the Thyroid dataset, features 21, 2, and 3 are consistently selected first by the model, while at feature 1 or 19, model suggestions start to diverge which feature to acquire next. Similarly in NHANES, features 2 and 30 form an initial decision baseline, before the model diverges into patient-specific decisions at features 33 or 45. In contrast, FACT and RL may change the feature acquisition order almost instantly, already at the first or second acquisition step, which may not always be justified or effective. In MNIST, FACT heatmaps show an outlining of the digit, as FACT multiplies the output of the de-noising auto-encoder with the feature-aggregation score. This strategy prioritizes high-intensity/-amplitude features, and leads to intuitive visualizations on MNIST, but does not directly translate to an efficient feature acquisition performance, as seen e.g. in the NHANES dataset. 
%In contrast, our AIG heatmap is visually dominated by the mean-feature baseline, following a strategy of imputation, in the absence of features during the peri-diagnostic acquisition process.
} Further, approaches like RL may choose features in random order (MNIST), or in order of least cost instead of relevance (Synthesized). In future work, we aim at investigating such phenomena from a medical perspective.

%%%%%%%%%%%%%%%%%%%%%%%%%%%%%%%%
% Feature acquisition heatmap
%%%%%%%%%%%%%%%%%%%%%%%%%%%%%%%%

%%%%%%%%%%%%%%%%%%%%%%%%%%%%%%%%%%%%%%%%%%%%%%%%%%%%%%%%%%%%%%%%%%%%%%%%%%%%%%%%%%%%%%%%
% Discussion
%%%%%%%%%%%%%%%%%%%%%%%%%%%%%%%%%%%%%%%%%%%%%%%%%%%%%%%%%%%%%%%%%%%%%%%%%%%%%%%%%%%%%%%%
%
\section{Conclusion}
\label{sec:conclusion}
We propose a novel method which can efficiently acquire features at test-time, through Accumulated Integrated Gradients (AIG) and network training with dropout at the input layer. We empirically show that our approach is cost- and feature-efficient when evaluated on two medical datasets and two explanatory toy datasets. Our proposed method enables patient-specific, peri-diagnostic decision support for clinicians, which could potentially optimize spending, maximize hospital resources, and reduce examination burden for patients. \toblue{Future work could address two important limitations of our work, which occur frequently in real-life clinical data, namely how to train a peri-diagnostic CADx system from data that is i) incomplete at training time and ii) made up of features from different modalities which are organized into blocks with acquisition costs that increase blockwise instead of one feature at a time.}\\
\\
\textbf{Acknowledgments:} This work was supported by the German Federal Ministry of Education and Health (BMBF) in connection with the foundation of the German Center for Vertigo and Balance Disorders (DSGZ) [grant number 01 EO 0901].

%%%%%%%%%%%%%%%%%%%%%%
% References
%%%%%%%%%%%%%%%%%%%%%%
\bibliographystyle{splncs04}
\bibliography{paper2729}

\end{document}